%% file: logicOptAI2submit.tex
\documentclass[graybox]{svmult}

\usepackage{type1cm}        
\usepackage{makeidx}         
\usepackage{graphicx}        
\usepackage{multicol}        
\usepackage[bottom]{footmisc}
\usepackage{comment}
\usepackage{amsmath}
\usepackage{amssymb}

\usepackage{newtxtext}       %
\usepackage[varvw]{newtxmath}       
\usepackage{bm}

\makeindex             

\input{extra_header_changes_Marty}

\newcommand{\half}{\frac{1}{2}}

\makeatletter
\newcommand{\widebarnew}[1]{%
	\begingroup
	\def\mathaccent##1##2{%
		\relax
		\ifmmode
		\macc@depth\@ne
		\let\macc@nucleus\@empty
		\setbox\z@\hbox{$\macc@style{#1}$}%
		\dimen@\wd\z@
		\advance\dimen@ -\scriptspace
		\divide\dimen@ 3
		\@tempdima\dimen@
		\advance\@tempdima -\macc@kerna
		\advance\@tempdima -\macc@kernb
		\macc@kern\@tempdima
		\let\mathaccent\save@mathaccent
		\fi
		\mathaccent"0362{#1}%
	}
	\overline{#1}%
	\endgroup
}
\makeatother

\begin{document}

\title*{Logic, Optimization, and Artificial Intelligence}
\author{J. N. Hooker\\May 2026}
\institute{Carnegie Mellon University, \email{jh38@andrew.cmu.edu}
}

%
%
\maketitle
1. Introduction\\
2. Probabilistic Logic \\
3. Belief Logics\\
4. Nonmonotonic and Many-valued Logics\\
5. Statistical Inference of Logical Formulas\\
6. Inference as Projection\\
7. Transparency through Postoptimality Analysis\\
8. Answer Set Programming Modulo Theories\\
9. Possible Research Directions\\
References\\

\bigskip

\abstract{Logic and optimization can, in combination, make valuable contributions to rule-based AI.  Logic is the obvious medium for encoding a rule base and drawing inferences from it, while optimization provides a powerful technology for computing inferences.  Their combination has taken on new relevance amid a growing concern for transparency in AI. 
which is important for reproducibility, explainability, trustworthiness, and fairness.  
Rule-based AI provides a natural solution to transparency that is becoming increasingly practical due to today's highly advanced optimization methods.  This article surveys several areas of logic-optimization partnership, including probabilistic logic, Bayesian logic, belief logics and Dempster-Shafer theory, nonmonotonic (default) logic, many-valued logics, and inference of logical formulas from noisy data based on Boolean regression.  It shows how to compute projections, the fundamental problem of both logic and optimization, using decision diagrams and logic-based Benders decomposition.  It describes the use of postoptimality analysis to explain how conclusions are reached, further enhancing transparency, as well as the role of optimization in answer set programming modulo theories.  The paper concludes by suggesting possible future research directions.}

\section{Introduction}
\label{sec:intro}
Logic and optimization have played an essential role in artificial intelligence (AI) since its inception.  The very first AI program (1956), {\em Logic Theorist} \cite{NewSim56}, proved theorems of logic and mathematics, including several from Russell and Whitehead's famous treatise {\em Principia Mathematica}.  The following year, the pathbreaking AI system {\em General Problem Solver} \cite{NewShaSim59} applied means-end analysis in a manner similar to logic programming.   As for optimization, the advent of machine learning and artificial neural networks immediately posed the problem of adjusting parameters to fit training data.  This optimization problem remains a central and challenging element of machine learning today. 

Logic and optimization are not only individually useful in AI, but they have much to offer when combined.  Together, they provide an ideal approach to implementing rule-based AI.  While connectionist machine learning has achieved impressive results, it increasingly poses the issue of transparency, which is important for reproducibility, explainability, trustworthiness, and fairness \cite{BurHub21,Hai20,Wal21,ZhoChe18}.  Rule-based AI provides a natural solution to transparency, because it allows knowledge to be explicitly encoded in rules from which relevant information can be inferred, and allows decisions to be deduced from a rule-based production system.  While neural networks provide a useful model of cognition, a rule-based framework has long served as an alternative model.  The well-known ACT-R framework inspired by AI pioneer Alan Newell, for example, places the agent's executive function in a rule-based production system \cite{AndBotByrDouLebQin04,LebAnd93,New94}.  

Logic and optimization are natural partners in a rule-based environment.  Logic provides the obvious framework for stating rules and deducing their implications, and even for their ethical assessment \cite{KimHooDon21}.  Optimization comes into play as a powerful method for computing logical inferences, as well as conceptual elucidation of the various logics used in AI.  Optimization solvers, particularly for linear and integer programming, not only deduce inferences but can provide a traceback of how they are derived, thus enhancing the natural transparency of rule-based AI.  These solvers have also improved enormously over the years.  For example, one study documents that integer programming solvers reduced solution times by a factor of 5 million between 1989 and 2024, quite apart from reductions due to faster machines \cite{Sol24}.  More fundamentally, we will see that optimization and logical inference are special cases of the same basic problem.

This paper presents a sampling of logic-optimization partnerships that have developed in parallel with machine learning and deserve ongoing consideration, especially in view of today's advanced optimization technology and concern for transparency.  We begin with probabilistic logic, 
which has a linear programming formulation that can be solved by modern column generation methods.  We also describe various extensions, including Bayesian logic, which combines probabilistic logic with Bayesian networks and makes use of nonlinear programming.  We examine belief logics as well, including Dempster-Shafer theory and variants that have easily-solved linear programming formulations.  We then take up two nonstandard logics that have seen application in AI, default logic (a variety of nonmonotonic reasoning) and many-valued logics.  Both benefit from integer programming models.  At this point we turn from deducing implications of logical formulas to inferring the formulas themselves from noisy data.  We describe a form of Boolean regression that can be calibrated by integer programming and measures statistical significance.  We then delve into the underlying connection between logic and optimization just mentioned by observing that both, at root, pose a projection problem.  We show how this problem can be addressed by applying a recently developed optimization method, logic-based Benders decomposition, to a knowledge base represented as a binary decision diagram (BDD).  We next specify three ways in which optimization methods can provide tracebacks that are potentially useful for enhancing transparency in AI.  One is classical linear programming sensitivity analysis, and the other two are inference-based and BDD-based postoptimality analysis for integer programming.  Following this, we combine default logic, logic-based Benders decomposition, and postoptimality analysis to illustrate the central role of optimization in answer set programming modulo theories.  We conclude by suggesting some possible avenues of future research.  

The emphasis throughout is on conveying the basic ideas and their motivation, rather than providing a complete technical description.  Details are available in the cited references, which cover both the methods discussed here and subsequent developments.

\section{Probabilistic Logic}

Although George Boole is best known for his work in propositional logic, he regarded probabilistic logic \cite{Boo1854} as his most important contribution.  It in fact displays stunning originality.  Although forgotten for more than a century, it is highly relevant to the project of artificial intelligence today.  It allows one to determine with what confidence one can draw inferences from propositions that are known only to be true with certain probabilities.  

Hailperin \cite{Hai76} observed in 1976 that Boole's probabilistic logic can be given a linear programming model.
A decade after Hailperin's work, Nilsson independently published a similar model in the AI literature \cite{Nil86}.  His paper gave rise to a number of subsequent contributions, many of which are surveyed in \cite{ChaHoo99,HanJau00,KliPar11}.

We begin by describing Boole's probabilistic logic and its linear programming model.  We show how the well-known technique of column generation can deal with the exponentially many variables in the model.  We then incorporate second-order probabilities without sacrificing linearity.  Conditional independence assumptions introduce nonlinear constraints, but we indicate how exploiting the structure of a Bayesian network can somewhat ameliorate the resulting computational challenge.

\subsection{The Basic Model} \label{sec:probBasic}

Probabilistic logic is best explained by an example \cite{ChaHoo99}.  Suppose the probabilities of three logical propositions are given as follows:
\begin{align}
& \mathrm{Pr}(x_1) = 0.9 \label{eq:prob1a} \\
& \mathrm{Pr}(x_1\supset x_2) = 0.8 \label{eq:prob1b}\\
& \mathrm{Pr}(x_2\supset x_3) = 0.7 \label{eq:prob1c} 
\end{align}
We are also given the conditional probability
\begin{equation}
\mathrm{Pr}(x_1\,|\, x_2,x_3) = 
\frac{\mathrm{Pr}(x_1,x_2,x_3)}{\mathrm{Pr}(x_2,x_3)} = 0.8 \label{eq:prob1d}
\end{equation}
Note that Pr$(x_1 \supset x_2)$ is not the conditional probability Pr$(x_2\,|\,x_1)$.  It is the probability of the material conditional $x_1\supset x_2$, which is equivalent to $\neg x_1\vee x_2$.  We wish to determine with what probability we can infer $x_3$.  Boole observed that we cannot deduce a precise probability for $x_3$, but we can deduce that its probability lies in a particular {\em range}.

In probabilistic logic, each possible assignment of truth values to ${\bm x} = (x_1,x_2,x_3)$ is regarded as a ``possible world'' (values 0 and 1 correspond to false and true).  Each possible world ${\bm x}$ has a probability $p_{\bm x}$, initially unknown.  For example, $p_{000}$ is the probability of the world in which $(x_1,x_2,x_3)=(0,0,0)$, and analogously for $p_{001}, \ldots, p_{111}$.  Then the probability 0.9 of proposition $x_1$, for example, is 
\[
p_{100}+p_{101}+p_{110}+p_{111}
\]
since $x_1$ is true in the corresponding possible worlds, and similarly for propositions $x_1\supset x_2$ and $x_2\supset x_3$.  The probability assignments \eqref{eq:prob1a}--\eqref{eq:prob1c} can be written as the linear equations
\begin{align}
& p_{100}+p_{101}+p_{110}+p_{111} = 0.9 \label{eq:prob2a} \\
& p_{000}+p_{001}+p_{010}+p_{011}+p_{110}+p_{111} = 0.8 \label{eq:prob2b} \\
& p_{000}+p_{001}+p_{011}+p_{100}+p_{101}+p_{111} = 0.7 \label{eq:prob2c} 
\end{align}
The conditional probability assignment \eqref{eq:prob1d} can be linearized as 
\[
\mathrm{Pr}(x_1,x_2,x_3) = 0.8\mathrm{Pr}(x_2,x_3)
\]
and therefore written
\begin{equation}
p_{111} - 0.8(p_{011}+p_{111}) = 0
\label{eq:prob2d}
\end{equation}

We can now deduce the range of possible values for the probability of $x_3$, which is equal to  
\begin{equation}
	p_{001}+p_{011}+p_{101}+p_{111}  \label{eq:prob3} 
\end{equation}
A sharp upper bound on this probability is the maximum of \eqref{eq:prob3} subject to the constraints \eqref{eq:prob2a}--\eqref{eq:prob2d} and the normalization and nonnegativity constraints
\begin{align}
& p_{000}+p_{001}+p_{010}+p_{011}+p_{100}+p_{101}+p_{110}+p_{111} = 1 \label{eq:prob2e} \\
& p_{000}, \ldots, p_{111} \geq 0 \label{eq:prob2f}
\end{align}
This is a linear programming problem with maximum value 0.7.  A sharp lower bound on the probability of $x_3$ is the minimum of \eqref{eq:prob3} subject to these constraints, which is 0.5.  The probability of $x_3$ must therefore be confined to the interval $[0.5,0.7]$.  Due to the convexity of the feasible set defined by the constraints, every probability in this interval is consistent with the assigned probabilities \eqref{eq:prob2a}--\eqref{eq:prob2d}.

In general, the probabilistic inference problem is
\begin{equation}
	\min/\max \; \big\{ 
	{\bm c}^{\!\intercal}{\! \bm p} \;\big| \; A{\bm p} = {\bm \pi}, \;
	B{\bm p} = {\bm 0}, \; {\bm 1}^{\intercal}{\!\bm p} = 1, \; {\bm p} \geq {\bm 0} \big\}
	\label{eq:prob4}
\end{equation}
where constraint $A{\bm p}={\bm \pi}$ specifies the categorical probabilities, $B{\bm p}={\bm 0}$ specifies the conditional probabilities, and ${\bm 1} = (1,1,\ldots, 1)$.  The probability assignments are unsatisfiable if \eqref{eq:prob4} has no feasible solution.  The example problem in this form is
\begin{equation}
\min/\max \left\{
[0\;1\;0\;1\;0\;1\;0\;1]{\bm p} \; \left| \;\;
\begin{array}{l}
   \left[
      \begin{array}{cccccccc}
         0&0&0&0&1&1&1&1 \\
         1&1&1&1&0&0&1&1 \\
         1&1&0&1&1&1&0&1 
       \end{array}
   \right]{\! \bm p} = 
   \left[
      \begin{array}{c}
         0.9 \\ 0.8 \\ 0.7
      \end{array}
   \right]
   \\[3.5ex]
   [0\;0\;0-\!0.8\;0\;0\;0\;0.2] {\bm p} = 0 \\[0.5ex]
   [1\;1\;1\;1\;1\;1\;1\;1]{\bm p}= 1, \;\;{\bm p} \geq {\bm 0}
\end{array}
\right.
\right\}
\end{equation}
where ${\bm p}=(p_{000}, \ldots, p_{111})$.  If desired, probability ranges rather than point values can be specified in \eqref{eq:prob4} while preserving the linear structure of the problem.  

\subsection{Column Generation}

A difficulty with problem \eqref{eq:prob4} is that it contains $2^n$ variables, where $n$ is the number of atomic propositions $x_i$.  One might think that this makes the problem computationally intractable, and Nilsson suggests as much in \cite{Nil93}.  However, the solution technique {\em column generation}, well known to the optimization community, is designed for just such cases.  It allows one to solve the problem by means of the simplex method while generating only a small fraction of the columns in matrices $A$ and $B$, and thereby using only a small fraction of the variables in ${\bm p}$.

Column generation is readily applied to probabilistic logic, as initially observed in \cite{Bru88,GeoKavPap88,Hoo88a,JauHanAra91,KavPap90}.  It begins with a restriction of \eqref{eq:prob4} in which $A$ and $B$ contain only a few columns.  It associates dual variables ${\bm u}$, ${\bm v}$, and $\alpha$ with the constraints $A{\bm p} = {\bm \pi}$, $B{\bm p} = {\bm 0}$, and ${\bm 1}^{\intercal}{\!\bm p} = 1$, respectively. These dual variables, whose values are calculated in the course of the simplex method, are used to compute the {\em reduced cost} of each variable $p_j$:
\begin{equation}
c_j - {\bm u}^{\intercal}A_j - {\bm v}^{\intercal}B_j - \alpha 
\label{eq:prob5}
\end{equation}
where $(c_j,A_j,B_j,1)$ is the column of $({\bm c}^{\intercal},A,B,{\bm 1}^{\intercal})$ corresponding to $p_j$.  Supposing for the moment that we are solving the minimization problem, each iteration of the simplex method selects a new column $(c_j,A_j,B_j,1)$ with a negative reduced cost for inclusion in the problem (when maximizing, a positive reduced cost is desired).  If no such column remains, an optimal solution has been found, at which point it is likely that only a small fraction of the columns in  $({\bm c}^{\intercal},A,B,{\bm 1}^{\intercal})$ have been selected.  

The problem of finding a $p_j$ with a negative reduced cost, known as the {\em pricing problem}, can be solved by minimizing \eqref{eq:prob5} over all columns $j$.  The nature of the pricing problem is best illustrated in the example.  A column of $({\bm c}^{\intercal},A,B,{\bm 1}^{\intercal})$ can be represented as $(y_0,y_1,y_2,y_3,z_1\!-\!0.8w_1,1)$, where the 0--1 variables $y_i$, $z_1$ and $w_1$ are defined by
\begin{equation}
\begin{array}{l}
y_0 \equiv x_3 \\
y_1 \equiv x_1 \\
y_2 \equiv (x_1\supset x_2) \\
y_3 \equiv (x_2\supset x_3) \\
z_1 \equiv (x_1\wedge x_2\wedge x_3) \\
w_1 \equiv (x_2\wedge x_3)
\end{array} \label{eq:prob7}
\end{equation}
The pricing problem can now be stated as minimizing
\begin{equation}
y_0 - (u_1y_1 + u_2y_2 + u_3y_3) - v_1(z_1-0.8w_1) - \alpha
\label{eq:prob8}
\end{equation}
subject to \eqref{eq:prob7}, which is a special case of the maximum satisfiability problem.  Solution of this problem yields a column with the smallest reduced cost, given by \eqref{eq:prob8}.  In practice, the problem is first solved by a fast heuristic method.  If this fails to yield a negative reduced cost, the problem is solved to optimality to confirm that no columns with a negative reduced cost remain.  The simplex method terminates when this is confirmed.

One way to solve the pricing problem to optimality is to formulate it as a 0--1 programming problem as proposed in \cite{Hoo88a}, which allows one to take advantage of powerful integer programming solvers.  The formulas in \eqref{eq:prob7} are converted to linear equations and inequalities, and \eqref{eq:prob5} then optimized subject to these constraints.  Conversion is straightforward.  For example, $y_0\equiv x_3$ becomes simply $y_0=x_3$, and the remaining formulas are expressed in conjunctive normal form before conversion to inequalities.  For instance, formula $y_2\equiv (x_1\supset x_2)$ becomes the conjunction of logical clauses
\[
\neg y_2\!\vee\!\neg x_1\!\vee\! x_2, \;\;\; y_2\!\vee\! x_1, \;\;\; y_2\!\vee\!\neg x_2
\]
which in turn become the three inequalities
\[
(\!1-\!y_2)+(1\!-\!x_1)+x_2\geq 1, \;\;\; y_2+x_1\geq 1, \;\;\; y_2 + (1\!-\!x_2) \geq 1  
\]

Another approach is to formulate the pricing problem as pseudo-Boolean optimization, as proposed in \cite{Bru88,JauHanAra91}.  In the example, this is accomplished by replacing variables $y_i$, $z_1$, and $w_1$ in the objective function \eqref{eq:prob8} with possibly nonlinear expressions in terms of the $x_i$s as follows:
\[
\begin{array}{l}
y_0 \leftarrow x_3 \\
y_1 \leftarrow x_1 \\
y_2 \leftarrow 1 - x_1x_2 \\
y_3 \leftarrow 1 - x_2x_3 \\
z_1 \leftarrow x_1x_2x_3 \\
w_1 \leftarrow x_2x_3 
\end{array}
\]
Then \eqref{eq:prob8} is minimized without constraints, using a heuristic method, or when a proof of optimality is required, using a pseudo-Boolean optimization method such as those described in \cite{BorHam02}.  

Subsequent developments in computational methods can be found in \cite{HanPer08,KliPar09,KliPar11}.  Some computationally easy cases of probabilistic logic are described in \cite{AndHoo97,AndPre01}.

\subsection{Extensions}

Problem \eqref{eq:prob4} is easily modified to derive the range of a conditional probability.  The desired conditional probability can be written ${\bm c}^{\!\intercal}{\!\bm p}/{\bm d}^{\intercal}{\!\bm p}$ for suitable coefficient vectors ${\bm c}$ and ${\bm d}$.  This results in the linear-fractional programming problem
\begin{equation}
\min/\max \; \Big\{ 
\frac{{\bm c}^{\!\intercal}{\! \bm p}}{{\bm d}^{\intercal}{\! \bm p}} \;\Big| \; A{\bm p} = {\bm \pi}, \;
B{\bm p} = {\bm 0}, \; {\bm 1}^{\intercal}{\!\bm p} = 1, \; {\bm p} \geq {\bm 0} \Big\}
\label{eq:prob9}
\end{equation}
A simple change of variable, and the introduction of a scalar variable $t$, transforms \eqref{eq:prob9} to a linear programming problem \cite{ChaCoo62}.  By setting ${\bm q}=t{\bm p}$, it is easily seen that \eqref{eq:prob9} is equivalent to 
\begin{equation}
\min/\max \; \big\{ 
{\bm c}^{\!\intercal}{\! \bm q} \;\big| \; A{\bm q} = t{\bm \pi}, \;
B{\bm q} = {\bm 0}, \; {\bm 1}^{\intercal}{\!\bm q} = t, \;{\bm d}^{\intercal}{\!\bm q}=1,\; {\bm q} \geq {\bm 0}, \; t\geq 0 \big\}
\label{eq:prob10}
\end{equation}
When a solution $({\bm q},t)$ of \eqref{eq:prob10} is found, the corresponding solution of \eqref{eq:prob9} is ${\bm p}={\bm q}/t$, with optimal value ${\bm c}^{\!\intercal}{\! \bm p}/{\bm d}^{\intercal}{\! \bm p}={\bm c}^{\!\intercal}{\! \bm q}$.  As an illustration, suppose we wish to find a probability range for $\mathrm{Pr}(x_3|x_1,x_2)$ in the above example.  In this case,
\[
\frac{{\bm c}^{\!\intercal}{\! \bm p}}{{\bm d}^{\intercal}{\! \bm p}} 
= \frac{[0\;0\;0\;0\;0\;0\;0\;1]{\bm p}}{[0\;0\;0\;0\;0\;0\;1\;1]{\bm p}}
\]
The deduced interval for $Pr(x_3|x_1,x_2)={\bm c}^{\!\intercal}{\! \bm q}$ is a point value, namely 4/7.

A second extension arises when the given probabilities $\pi_i$ are intervals rather than point values.  Then the inference problem \eqref{eq:prob4} is
\begin{equation}
\min/\max \; \big\{ 
{\bm c}^{\!\intercal}{\! \bm p} \;\big| \; {\bm \pi}^{\mathrm{lo}} \leq A{\bm p} \leq {\bm \pi}^{\mathrm{hi}}, \; B{\bm p}={\bm 0},\; {\bm 1}^{\intercal}{\!\bm p} = 1, \; {\bm p} \geq {\bm 0} \big\}
\label{eq:prob11}
\end{equation}
A probability range for ${\bm c}^{\!\intercal}{\! \bm p}$ can be derived by linear programming as before, but it might be regarded as too conservative.  Probabilities at the ends of a range $[\pi^{\mathrm{lo}}_i,\pi^{\mathrm{hi}}_i]$ assigned to a premise are generally less likely to be the true probability than those near the middle.  Probabilities at the ends of the derived range may therefore be extremely unlikely, because they typically occur only when a number of the given probabilities are near the ends of their intervals.  The derived range can therefore be misleadingly wide.  

One possible remedy for this problem is to derive a ``most likely'' probability for ${\bm c}^{\!\intercal}{\! \bm p}$ by computing a maximum entropy solution \cite{Kan89}.  Yet this poses a difficult nonlinear optimization problem \cite{HanJau00}, and it is equally misleading, because the true probability of the inferred proposition may be quite far from its most likely value.  Another remedy, described in \cite{ChaHoo99}, is to specify a second-order probability distribution over each probability interval $[\pi^{\mathrm{lo}}_i, \pi_i^{\mathrm{hi}}]$ to indicate that probabilities near the middle of the interval are more likely.  To preserve the linearity of the optimization problem, we can suppose the logarithm of the second-order probability density function is piecewise linear and therefore the lower envelope of a set of lines:
\[
\log \mathrm{Pr}(\pi_i=\rho) = \min_{k=1,\ldots,k_i} \!\{\alpha_{ik}\rho + \beta_{ik}\}
\]
Let $A^i$ be row $i$ of $A$, which corresponds to proposition $i$.  Assuming independence of the second order probabilities, the logarithm of the joint probability that the $m$ propositions have probabilities $A^1{\bm p},\ldots,A^m{\bm p}$ is
\begin{equation}
\sum_{i=1}^m \min_{k=1,\ldots,k_i}\!\{\alpha_{ik}A^i{\bm p} + \beta_{ik}\}
\label{eq:prob20}
\end{equation}
To eliminate solutions that are extremely unlikely, one can impose a lower bound $L$ on \eqref{eq:prob20}.  The probabilistic inference problem becomes the linear programming problem
\[
\min/\max \; \left\{ 
{\bm c}^{\!\intercal}{\! \bm p} \;\left| 
\begin{array}{l}\sum_{i=1}^m \ell_i \geq L, \;
\ell_i \leq \alpha_{ik}A^i{\bm p} + \beta_{ik}, \;k=1,\ldots, k_i, \; i=1,\ldots,m \\[0.7ex]
{\bm \pi}^{\mathrm{lo}} \leq A{\bm p} \leq {\bm \pi}^{\mathrm{hi}}, \; B{\bm p}={\bm 0},\; {\bm 1}^{\intercal}{\!\bm p} = 1, \; {\bm p} \geq {\bm 0} 
\end{array}
\right.
\right\}
\]
which can again be solved by column generation.  The probability range can now be reduced to a more realistic interval by increasing $L$.  Second-order distributions of the conditional probabilities can be similarly accommodated.  A similar model in \cite{AndHoo96} shows how to assign probabilities point values that are uncertain, where the degree of uncertainty is based on information from several sources.

\subsection{Bayesian Logic} \label{sec:BayesianLogic}

One of our most valuable sources of probabilistic knowledge is the independence of most events.  It is convenient, for example, that the probability of an earthquake in Asia is independent of the weather in North America.  Ironically, the independence relations that make things easier to understand also make a probabilistic logic model harder to solve, because they introduce nonlinearities.  
Despite this, it would be desirable to incorporate independence assumptions into Boole's probabilistic logic.  {\em Bayesian logic} accomplishes this by identifying the nodes of a Bayesian network with logical formulas \cite{AndHoo94}, thereby taking advantage of the elegant way in which Bayesian networks represent conditional independence.  

\begin{figure}[!h]
	\centering
	\includegraphics[scale=1.25,clip=true,trim=0 5 0 0]{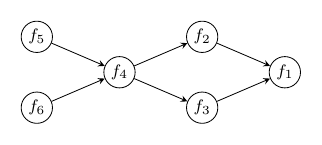} \hspace{2ex}
	\caption{A simple Bayesian network.}
	\label{fig:bayes1}
\end{figure}

Figure~\ref{fig:bayes1} represents a small Bayesian network for formulas $f_1,\ldots,f_6$.  We can suppose that probabilities are specified (or bounded) for some or all of the nodes, conditioned on the node's immediate predecessors, as in the conditional probability Pr$(f_1|f_2,f_3)$.  Marginal probabilities can be specified (or bounded) for nodes without predecessors, such as $f_5$ and $f_6$.  The formulas $f_i$ contain propositional variables $x_j$ as in probabilistic logic.  The formulas are therefore related logically as well as through conditional probabilities.  

The essence of a Bayesian network is a Markovian property according to which conditional probabilities depend only on immediate parent nodes.  Thus in Fig.~\ref{fig:bayes1}, Pr$(f_1|f_2,f_3,f_4)=\mathrm{Pr}(f_1|f_2,f_3)$.  Consequently, a set of nodes with common parents are conditionally independent when their probabilities are conditioned on their parents.  For example, $f_2$ and $f_3$ are conditionally independent given $f_4$, so that Pr$(f_2|f_3,f_4)=\mathrm{Pr}(f_2|f_4)$.  Since $f_4$ is a proposition, one may wish to suppose that $f_2$ and $f_3$ are independent given $\neg f_4$ as well as given $f_4$.  We will indicate this by writing Pr$(f_2|f_3F_4)=\mathrm{Pr}(f_2|F_4)$, where $F_4$ varies over $f_4$ and $\neg f_4$.  

An optimization model is obtained by adding network-encoded nonlinear independence conditions to the linear programming model for probabilistic logic.  For example, the independence assumption Pr$(f_2|f_3F_4)=\mathrm{Pr}(f_2|F_4)$ is captured by the constraints 
\begin{equation}
\mathrm{Pr}(f_2,f_3,F_4)\mathrm{Pr}(F_4) = \mathrm{Pr}(f_2,F_4)\mathrm{Pr}(f_3,F_4),
\;\;\mbox{for}\; F_4=f_4,\neg f_4
\end{equation}
where Pr$(f_2,f_3,F_4)$ is the joint probability of $f_2$, $f_3$, and $F_4$.    Such constraints are not only nonlinear but can grow exponentially in number.  If the probabilities are conditioned on several formulas, one must write constraints for all true-false combinations of these formulas. 

\begin{figure}[!h]
	\centering
	\includegraphics[scale=1.1,clip=true,trim=0 10 0 0]{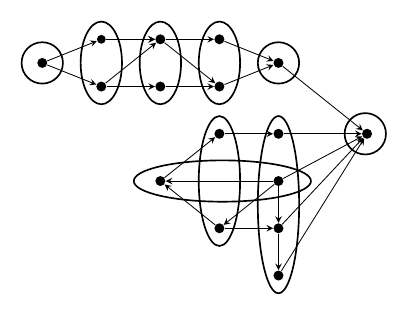} \hspace{2ex}
	\caption{Ancestral sets of the rightmost node.}
	\label{fig:bayes3}
\end{figure}

\begin{figure}[!h]
	\centering
	\includegraphics[scale=1.1,clip=true,trim=0 10 0 0]{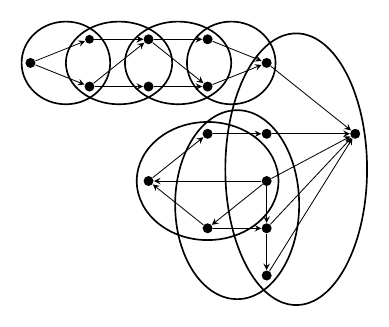} \hspace{2ex}
	\caption{Extended ancestral sets of the rightmost node.}
	\label{fig:bayes4}
\end{figure}

Fortunately, it is shown in \cite{AndHoo94,ChaHoo99} that the number of constraints is substantially limited in many networks. The idea is illustrated in Figs.~\ref{fig:bayes3} and~\ref{fig:bayes4}.  To compute bounds on the probability of the rightmost node, the number of independence constraints is at worst exponential in the size of the largest {\em extended ancestral set} of that node.  Figure~\ref{fig:bayes3} shows the ancestral sets, and Fig.~\ref{fig:bayes4} the extended ancestral sets.  The ancestral sets are defined recursively in terms of parent sets.  Given a set $S$ of nodes, the parent sets of $S$ are obtained by identifying the set $S'$ of nodes outside $S$ that are immediate predecessors of some node in $S$, and then splitting $S'$ into maximal dependent subsets.  The ancestral sets of a node are its parent sets, the parent sets of its parent sets and so forth.  An extended ancestral set consists of an ancestral set augmented by the nodes in its parent sets.

A variation of Bayesian logic was recently proposed for a {\em logical credal network} \cite{MarQiaGraBhaBarGaoRieSah22}.  Credal networks are modified Bayesian networks in which probabilities are specified in various imprecise ways, generalizing the intervallic specifications used here.  A logical credal network associates logical formulas with nodes as does Bayesian logic.  In \cite{MarQiaGraBhaBarGaoRieSah22}, the Markovian property of conventional Bayesian networks is replaced by a modified property that permits the network to contain cycles.  Exact and heuristic solution methods for computing probability bounds in logical credal networks are described in \cite{CozMarLeeGraRieBha24,MarLeeBhaCozGra24,MarQiaGraBhaBarGaoRieSah22}.  Credal networks in general, including optimization methods for solving them, are comprehensively surveyed in \cite{MauCoz20}.  Integer programming methods are described in \cite{MaudeCamBenZaf12}.

\section{Belief Logics}

Like probabilistic logic, a logic of belief functions distributes probability mass over subsets of possible worlds.  It differs in that the probability mass indicates the degree of evidence for, or credibility of, propositions rather than their  probability in a classical sense.  It might therefore be referred to as evidence mass.  A further difference is that, whereas assigning a specific probability mass to a set of possible worlds {\em fixes} the total probability of those worlds, assigning a specific evidence mass {\em adds to} the total evidence for those worlds.  

We begin below with a basic belief logic and its linear programming formulation.  Unlike probabilistic logic, it does not contain exponentially many variables and therefore has no need of column generation for its solution.  
We then describe how Dempster-Shafer theory augments this basic model with a rule for combining possibly contradictory evidence sources (Dempster's combination rule).  Following this, we drop the nonlinear independence assumption of Dempster-Shafer theory, while retaining Dempster's combination rule, to permit a linear programming formulation.

\subsection{Logic of Belief Functions} \label{sec:beliefFunctions}

Belief functions can be illustrated by a small example, taken from \cite{AndHoo94}.  Suppose we have propositions $x_1$, $x_2$, and $x_1\!\wedge\! x_2$, which correspond to possible world sets $S_1=\{(1,0),(1,1)\}$, $S_2=\{(0,1),(1,1)\}$, and $S_0=\{(1,1)\}$, respectively.  We have some evidence for $x_1$ and $x_2$ and wish to determine the implied evidential support for $x_1\!\wedge\! x_2$.  Let $m(S_1)$ be the fraction of available evidence that specifically supports the proposition $x_1$ corresponding to set $S_1$, and similarly for $m(S_2)$ and $m(S_0)$.  We also suppose that $m(\Theta)$ is the fraction of evidence that supports no particular proposition, where $\Theta$ is the set of all possible worlds.  Dempster and Shafer refer to $m(S_i)$ as a ``basic probability number,'' but we will refer to it as an {\em evidence allotment}.  We must have $m(\Theta) + \sum_{i}m(S_i)=1$.

The accumulated evidence for $x_1$ consists of the evidence that supports $x_1$ and any propositions that imply $x_1$, in this case $x_1\!\wedge\! x_2$.  Thus the total support for set $S_1$ is {\it Bel}$(S_1)=m(S_0)+m(S_1)$, where {\it Bel} is a {\em belief function}.  The belief values {\it Bel}$(S_2)$ and {\it Bel}$(S_0)$ are similarly defined.  If we are given that {\it Bel}$(S_1)=0.9$ and {\it Bel}$(S_2)=0.8$, we have
\begin{equation}
\begin{array}{l}
m(S_0)+m(S_1) = 0.9 \\
m(S_0)+m(S_2) = 0.8 \\
m(S_0)+m(S_1)+m(S_2)+m(\Theta) = 1 \\
m(S_i)\geq 0, \;\mbox{all}\;i
\end{array} \label{eq:bel1}
\end{equation}
The evidence allotments $m(S_i)$ must satisfy these constraints.  Since evidence is allocated to no proper subset of $S_0$, the evidential support for $x_1\!\wedge\! x_2$ is {\it Bel}$(S_0)=m(S_0)$.  This support level must fall in the range obtained by maximizing and minimizing $m(S_0)$ subject to \eqref{eq:bel1}, which yields the interval $[0.7,0.8]$.

In general, suppose we are given belief values {\it Bel}$(S_i)$ for $i=1,\ldots,m$.  The linear programming model is
\[
\min/\max\;
\left\{ \sum_{S\subseteq S_0} \!m(S) \; \left| \;
\begin{array}{l}
{\displaystyle
\sum_{S\subseteq S_i}\! m(S) = \mathrm{\it Bel}(S_i), \;i=1,\ldots, m 
} \\[-1ex]
{\displaystyle
m(\Theta)+\sum_{i=1}^m m(S_i) = 1
} \\[3ex]
m(\Theta)\geq 0, \;m(S_i)\geq 0, \;i=1,\ldots,m
\end{array}
\right.
\right\}
\]
Note that possible worlds do not play an explicit role in this model.  The model therefore grows only linearly with the number of propositions assigned belief values and can therefore be solved very rapidly.  In practice, set inclusion \mbox{$S\!\subseteq\! S_i$} can be interpreted as logical entailment between the propositions corresponding to $S$ and~$S_i$.

\subsection{Dempster-Shafer Theory}

The belief logic of the previous section supposes that all available evidence has been pooled, so that evidence allotments $m(S)$ can be made on that basis.  It also recognizes that the belief level of a given proposition may be underdetermined by available evidence.  Dempster-Shafer theory \cite{Dem67,Dem68,Sha76} addresses the issue of how to combine different sources of evidence to obtain a single evidence allotment function $m(\cdot)$.  It accomplishes this in such a way as to obtain point values for the function $m(\cdot)$, and therefore for any belief value {\it Bel}$(S_0)$ one wishes to deduce from the evidence base.  The method rests on rather strong independence and normalization assumptions.  However, one can drop the independence assumption and deduce intervals for $m(\cdot)$, rather than point values, by means of linear-fractional programming.

We again illustrate with an example.  As before, the propositions of interest are $x_1$, $x_2$, and $x_1\!\wedge x_2$, corresponding to sets $S_1,S_2$ and $S_0$, and we wish to deduce the level of evidence for $x_1\!\wedge x_2$.  Suppose that evidence source~1 determines an evidence allotment function $m_1(\cdot)$ by specifying $m_1(S_1)=0.9$ and $m_1(\Theta)=0.1$.  Evidence source~2 specifies $m_2(S_2)=0.8$ and $m_2(\Theta)=0.2$.  Then {\em Dempter's combination rule} derives a combined allotment function $\widehat{m}(\cdot)=m_1(\cdot)\otimes m_2(\cdot)$ by taking cross-products of evidence allotments to intersecting sets as shown in Table~\ref{ta:bel1}.  This operation assumes that evidence sources are independent in some sense.  Since none of the intersections are empty, we can set $m(\cdot)=\widehat{m}(\cdot)$; empty intersections require renormalization, as explained below.  Because $S_0$ contains neither  of the other sets, its belief value is simply {\it Bel}$(S_0)=m(S_0)=0.72$.  Note that this point value lies within the interval $[0.7,0.8]$ that was derived for {\it Bel}$(S_0)$ in the previous section.  

\begin{table}[t!]
\centering
\caption{Example of Dempster's combination rule that does not require normalization.} \label{ta:bel1}
\begin{tabular}{r@{\hspace{1ex}}|c|c|}
\multicolumn{1}{r}{} & \multicolumn{1}{c} {$m_2(S_2)=0.8$} & \multicolumn{1}{c}{$m_2(\Theta)=0.2$} \\
	           \cline{2-3}
 & & \\[-1.8ex]
$m_1(S_1)=0.9$ & \hspace{1ex}
                 $\begin{array}{c} 
	              \widehat{m}(S_1\cap S_2)= \\
	              \widehat{m}(S_0)=0.72 
	              \end{array}$
	             \hspace{1ex}
	           & \hspace{1ex}
	              $\begin{array}{c}
	              \widehat{m}(S_1\cap\Theta)= \\
	              \widehat{m}(S_1) = 0.18
	              \end{array}$
	              \hspace{1ex}
\\[1.5ex]
\cline{2-3}
 & & \\[-1.8ex]
$m_1(\Theta)=0.1$ & $\begin{array}{c} 
	\widehat{m}(\Theta\cap S_2)= \\
	\widehat{m}(S_2)=0.08 
\end{array}$
& $\begin{array}{c}
	\widehat{m}(\Theta\cap\Theta)= \\
	\widehat{m}(\Theta) = 0.02
\end{array}$
\\[1.5ex]
\cline{2-3}
\end{tabular}
\end{table}

In general, we have
\[
\widehat{m}(S) = m_1(S)\otimes m_2(S) = \hspace{-1ex} \sum_{\substack{A,B\\S=A\cap B}} \hspace{-1.3ex} m_1(A)m_2(B)
\]
If there are three evidence sources, one can compute $\widehat{m}(S)$ recursively as $\widehat{m}(S)=m_1(S)\otimes (m_2(S)\otimes m_3(S))$, and similarly for more numerous sources.  If one or more intersections $A\cap B$ are empty, we must renormalize the evidence allotments.  For example, the allotment functions shown in Table~\ref{ta:bel2} assign positive evidence to contradictory propositions, since $\widebarnew{S}_1$ is the complement of $S_1$.  Since there can be no evidence for a logical contradiction, we assign $m(S_1\!\cap\widebarnew{S}_1)=m(\emptyset)=0$.  We now renormalize the remaining evidence allotments, yielding $m(S_1)=\frac{1}{3}$, $m(\widebarnew{S}_1)=\frac{1}{6}$, and $m(\Theta)=\frac{1}{2}$.  More generally,
\[
m(S) = \left\{
\begin{array}{ll}
{\displaystyle \frac{\widehat{m}(S)}{1-\widehat{m}(\emptyset)}, 
} & \mbox{if}\; S\neq \emptyset \\[2ex]
0, & \mbox{if}\; S=\emptyset 
\end{array}
\right.
\]
In this manner, Dempster's combination rule can reconcile conflicting evidence sources, although 
one might question whether renormalization is a justifiable method for doing so.  Finally, 
the belief value of $S_0$ is simply
\[
\mathrm{\it Bel}(S_0) = \sum_{S\subseteq S_0} \hspace{-0.5ex}m(S)
\]

\begin{table}[!t]
\centering
\caption{Example of Dempster's combination rule that requires normalization.} \label{ta:bel2}
\begin{tabular}{r@{\hspace{1ex}}|c|c|}
\multicolumn{1}{r}{} & \multicolumn{1}{c} {$m_2(\widebarnew{S}_1)=0.25$} & \multicolumn{1}{c}{$m_2(\Theta)=0.75$} \\
\cline{2-3}
& & \\[-1.8ex]
$m_1(S_1)=0.4$ & \hspace{1ex}
           		$\begin{array}{c} 
				\widehat{m}(S_1\cap \widebarnew{S}_1)= \\
				\widehat{m}(\emptyset)=0.1 
				\end{array}$
				\hspace{1ex}
				& \hspace{1ex}
				$\begin{array}{c}
				\widehat{m}(S_1\cap\Theta)= \\
				\widehat{m}(S_1) = 0.3
				\end{array}$
				\hspace{1ex}
\\[1.5ex]
\cline{2-3}
& & \\[-1.8ex]
$m_1(\Theta)=0.6$ & $\begin{array}{c} 
				\widehat{m}(\Theta\cap \widebarnew{S}_1)= \\
				\widehat{m}(\widebarnew{S}_1)=0.15
				\end{array}$
				& $\begin{array}{c}
				\widehat{m}(\Theta\cap\Theta)= \\
				\widehat{m}(\Theta) = 0.45
				\end{array}$
\\[1.5ex]
\cline{2-3}
\end{tabular}
\end{table}

\subsection{Belief Logic with Dempster's Combination Rule}

The independence assumption in Dempster-Shafer theory can be dropped if we are content with inferred intervals rather than point values for the evidence allotment function $m(\cdot)$.  This yields a belief logic similar to that described in Section~\ref{sec:beliefFunctions}, but with a mechanism for combining evidence sources.  It also permits a linear programming model.

In the example of Table~\ref{ta:bel1}, we observe that
\begin{equation}
\begin{array}{l}
\widehat{m}(S_1\!\cap\! S_2) + \widehat{m}(S_1) = 0.9 \\[0.3ex]
\widehat{m}(S_2) + \widehat{m}(\Theta) = 0.1 \\[0.3ex]
\widehat{m}(S_1\!\cap\! S_2) + \widehat{m}(S_2) = 0.8 \\[0.3ex]
\widehat{m}(S_1) + \widehat{m}(\Theta) = 0.2 \\[0.3ex]
\widehat{m}(S_1), \widehat{m}(S_2), \widehat{m}(S_1\!\cap\! S_2), \widehat{m}(\Theta) \geq 0
\end{array} \label{eq:bel10}
\end{equation}
where any one of the four equations is redundant of the others.  By minimizing and maximizing $\widehat{m}(S_1\!\cap\! S_2)$ subject to \eqref{eq:bel10}, we obtain the range $[0.7,0.8]$ for $\widehat{m}(S_1\!\cap\! S_2)$, which contains the point value 0.72 obtained under an independence assumption.  A similar approach works when renormalization is necessary.  From Table~\ref{ta:bel2}, we have 
\begin{equation}
\begin{array}{l}
\widehat{m}(\emptyset) + \widehat{m}(S_1) = 0.4 \\[0.3ex]
\widehat{m}(\widebarnew{S}_1) + \widehat{m}(\Theta) = 0.6 \\[0.3ex]
\widehat{m}(\emptyset) + \widehat{m}(\widebarnew{S}_1) = 0.25 \\[0.3ex]
\widehat{m}(S_1) + \widehat{m}(\Theta) = 0.75 \\[0.3ex]
\widehat{m}(S_1), \widehat{m}(\widebarnew{S}_1), \widehat{m}(\emptyset), \widehat{m}(\Theta) \geq 0
\end{array} \label{eq:bel11}
\end{equation}
Since $m(S_1)= \widehat{m}(S_1)/(1-\widehat{m}(\emptyset))$, we can obtain a range for $m(S_1)$ by minimizing and maximizing this fraction subject to \eqref{eq:bel11}.  This can be accomplished by fractional-linear programming, yielding the interval $[0.2,0.4]$, which contains the point value $m(S_1)=\frac{1}{3}$ derived above.

In general, a range for $m(S_0)$, given nonempty $S_0$, is found as follows.  Suppose that evidence source~1 provides positive evidence mass for sets $S_1,\ldots,S_m$, and evidence source~2 provides positive mass for $T_1,\ldots,T_n$.  Then we wish to solve
\[
\min/\max \;
\left\{
\frac{\widehat{m}(S_0)}{1-\widehat{m}(\emptyset)} \;
\left| \;
\begin{array}{l}
\sum_{j=1}^n \widehat{m}(S_i\!\cap\! T_j) = m_1(S_i), \; i=1,\ldots, m \\[1ex]
\sum_{i=1}^m \widehat{m}(S_i\!\cap\! T_j) = m_2(T_j), \; j=1,\ldots, n \\[1ex]
\widehat{m}(\cdot)\geq 0
\end{array}
\right.
\right\}
\]
This becomes a linear programming problem after the variable change $\mu(\cdot)=\widehat{m}(\cdot)t$:
\[
\min/\max \;
\left\{
\mu(S_0) \;
\left| \;
\begin{array}{l}
\sum_{j=1}^n \mu(S_i\!\cap\! T_j) = m_1(S_i)t, \; i=1,\ldots, m \\[1ex]
\sum_{i=1}^m \mu(S_i\!\cap\! T_j) = m_2(T_j)t, \; j=1,\ldots, m \\[1ex]
t - \mu(\emptyset) = 1; \;\; \mu(\cdot), t\geq 0
\end{array}
\right.
\right\}
\]

When there are a large number of evidence sources, recursive computation of $m(S_0)$ 
can become difficult.  The number of set intersections, and thus the number of sets with positive mass, can increase exponentially with the number of sources.  Methods for simplifying the computations, using a set covering optimization model, are discussed in \cite{ChaHoo99,DugSan94}.  In any event, it is likely that only a handful of evidence sources are relevant to a given conclusion in a given application.  

A survey of algorithms for Dempster-Shafer theory can be found in \cite{Wil00}, and a general discussion of semantics for belief logics in \cite{HaeRomWheWil08}.  Other mathematical programming embeddings of logic, including predicate logic, are presented in \cite{BorChaMit02,ChaHoo99}.

\section{Nonmonotonic and Many-Valued Logics}

This section deals with two nonstandard logics that have played a role in artificial intelligence, nonmonotonic logic and many-valued logic.  It shows how optimization can make substantial contributions to their implementation, and even lend insight into the underlying ideas.  

In {\em nonmonotonic logic}, the addition of new facts to a knowledge base may require the withdrawal of previously inferred conclusions.  We show how integer programming can capture the semantics of default logic (a popular variety of nonmonotonic logic) and compute satisfying solutions.

Many-valued logic can assign truth values between true and false, values that may either be discrete or lie on a continuous scale.  This permits a knowledge base to indicate the degree of confidence with which propositions can be asserted.  Mixed integer/linear programming (MILP) is well suited to determine truth values of propositions deduced from the knowledge base.

\subsection{Default Logic} \label{sec:default}

In {\em default logic}, propositions with an ``unless'' clause are enforced only if the unless clause is not known to be true.  This type of logic is closely related to answer set programming and negation-as-failure in logic programming.  A key concept in the semantics of default logic is that of a {\em stable model} \cite{GelLif88,MarNerRem94}.  Optimization can be helpful in this context because all stable models can be identified with the assistance of integer programming \cite{BelNerNgSub94,ChaHoo99}.

Default logic is normally applied to a set of guarded rules having the form on the left:
\[
P\!\rightarrow\! x_i\; \mbox{unless}\;G, \;\;\;\mbox{or}\;\;\; \neg P\vee\!x_i  \vee G
\]
where $P$ is a conjunction of zero or more atoms (atomic propositions) and logical formula $G$ is the guard.
The rule is semantically equivalent to the formula on the right.  One is permitted to use the rule $P\!\rightarrow\! x_i$ in a deduction as long as $G$ is not known to be true.  

In this context, a {\em model} for a set of guarded rules is a collection of atoms that, when set to true, satisfy the rules when the remaining atoms are presumed false.  The concept of stable model can be explained with the following example.  Consider the guarded rules on the left below, which are semantically equivalent to the formulas on the right.
\begin{equation}
\begin{array}{l}
x_1\!\rightarrow\! x_2 \;\mbox{unless}\; x_5 \\
x_1\!\rightarrow\!x_3 \;\mbox{unless}\; (x_2\!\vee\!x_4) \\
(x_1\!\wedge\! x_4)\!\rightarrow\! x_5 \\
\rightarrow\! x_1 \;\mbox{unless}\; (x_4\wedge x_5)
\end{array} \hspace{5ex}
\begin{array}{l}
\neg x_1\!\vee\! x_2\!\vee\! x_5 \\
\neg x_1\!\vee\! x_3\!\vee\! x_2\!\vee\! x_4 \\
\neg x_1\!\vee\! \neg x_4\!\vee\! x_5 \\
x_1\!\vee\! (x_4\!\wedge\! x_5)
\end{array} \label{eq:mono10}
\end{equation}
Note that a guard can be empty, in which case the rule is always enforced.  A given model allows rules to be used when it fails to activate their guards.  For example, the model $\{x_1.x_2\}$ admits the rules 
\begin{equation}
\begin{array}{l}
x_1\!\rightarrow\! x_2 \\
(x_1\!\wedge\!x_4)\! \rightarrow\! x_5 \\
\rightarrow\! x_1
\end{array} \label{eq:mono11}
\end{equation}
where \eqref{eq:mono11} is known as the {\em Gelfond-Lifschitz transform} of \eqref{eq:mono10} with respect to the given model \cite{GelLif88}.  A {\em stable} model is one that minimally satisfies its Gelfond-Lifschitz (G-F) transform; that is, flipping any atom from true to false would no longer satisfy the transform.  It is easily checked that $\{x_1,x_2\}$ is a stable model.  

Integer programming can identify all stable models by systematically finding all minimal models of the original rule base and checking which ones are minimal for the corresponding Gelfond-Lifschitz transform.  A minimal model for \eqref{eq:mono10}, for example, is found by minimizing $\sum_i x_i$ subject to linear inequalities that encode the logical formulas in \eqref{eq:mono10}:
\begin{equation}
\min\; \left\{
\sum_i x_i \; \left| \;
\begin{array}{l}
(1\!-\!x_1)+x_2+x_3 \geq 1 \\
(1\!-\!x_1)+x_2+x_3+x_4 \geq 1 \\
(1\!-\!x_1)+(1\!-\!x_4)+x_5 \geq 1 \\
x_1+x_4 \geq 1, \;\; x_1+x_5\geq 1 \\
x_i\in\{0,1\}, \;\mbox{all}\; i
\end{array}
\right.
\right\}
\label{eq:mono12}
\end{equation}
Rules must be converted to conjunctive normal form before encoding, as illustrated by the last rule in \eqref{eq:mono10}, which is converted to $(x_1\!\vee\!x_4)\wedge (x_1\!\vee\!x_5)$.  One optimal solution of \eqref{eq:mono12} is $(x_1,\ldots,x_5)=(1,1,0,0,0)$, which corresponds to the model $\{x_1,x_2\}$, already found to be stable.  Another minimal model can be found by adding the constraint $x_1+x_2\leq 1$ to exclude the minimal model $\{x_1,x_2\}$, whereupon solution of \eqref{eq:mono12} yields $(x_1,\ldots,x_5)=(0,0,0,1,1)$.  This is not minimal in the G--L transform, which consists of the single rule \mbox{$(x_1\!\wedge\! x_4)\!\rightarrow \!x_5$}.  In fact, this rule is satisfied by flipping both ones to zero.  Adding the further constraint \mbox{$x_4\!+\! x_5\leq 1$} yields a minimal model $\{x_1,x_3,x_5\}$, which is not minimal in the corresponding G--L transform.  Finally, adding $x_1\!+\! x_3\!+\! x_5\leq 2$ makes \eqref{eq:mono12} infeasible.  Thus \eqref{eq:mono10} has no more minimal models, and $\{x_1,x_2\}$ is its only stable model.

Because the G--L transform is always a Horn system, unit propagation can check for minimality, with no need for integer programming at this stage. A Horn system consists of rules of the form $P\!\rightarrow x_i$, where $P$ is a conjunction of zero or more atoms, and $x_i$ may be absent (in which case $\neg P$ is asserted).  There is exactly one minimal model (possibly empty) for a Horn system, which is found by applying unit propagation and noting which atoms are fixed to true.  These atoms comprise the minimal model.  Unit propagation proceeds by finding a rule in which $P$ is empty, fixing its consequent $x_i$ to true, and removing $x_i$ from all other antecedents $P$ in the rule base.  This is repeated until no empty antecedents remain.  For example, unit propagation applied to \eqref{eq:mono11} first fixes $x_1$ to true, and then $x_2$, whereupon the rule $x_4\!\rightarrow\! x_5$ remains.  Thus $\{x_1,x_2\}$ is the unique minimal model of \eqref{eq:mono11}.  Horn systems can also be solved by linear programming and have a number of interesting polyhedral properties \cite{ChaHoo91,JerWan90}.

Subsequent research on integer programming methods for answer set programming and stable semantics includes \cite{Ben05,LiuJanNie12,LiuJanNie14,OsoDiaSan17}.

\subsection{Many-valued Logic} \label{sec:manyvalued}

Truth values in many-valued logic are frequently taken to be equally spaced numbers in the interval $[0,1]$, with larger values corresponding to greater degree of confidence.  
If desired, all real values in the interval $[0,1]$ can be regarded as truth values.  Logical operators and connectives are semantically defined as functions of these truth values.  For example, the truth value of $\neg P$ is $v(\neg P)=1-v(P)$, where $v(P)$ is the truth value of $P$.  The truth value of the conditional $P\supset Q$ is frequently defined to be
\begin{equation}
	v(P\supset Q)=\min\{1, 1-v(P)+v(Q)\} \label{eq:multi1}
\end{equation} 

Mixed integer/linear programming (MILP) provides a natural  mechanism for determining what truth values can inferred for a given formula, given premises with specified truth values \cite{Hah94}.  MILP presupposes that the operators and connectives are {\em MILP-representable}, but a necessary and sufficient condition for representability is known, easily checked, and rather weak. 
In particular, Jeroslow \cite{Jer87} showed that an optimization problem is (finitely) MILP representable if and only if its feasible set\footnote{Technically, this condition applies to the epigraph rather than the feasible set, but the distinction is unnecessary here.}  is the union of finitely many polyhedra that have the same recession directions.  These are directions in which one can go forever without leaving the set.  Thus ${\bm d}$ is a recession direction for set $S$ if for some ${\bm v}\in S$, ${\bm v}+\alpha {\bm d}\in S$ for all $\alpha\geq 0$.   

In classical logic, the truth value of a formula $P$ is indicated by asserting $P$ or $\neg P$.  In multi-valued logic, one can specify (or bound) the truth value of $P$ by means of a more general {\em signed formula}.\footnote{We slightly modify the notation in \cite{Hah94}}  For example, the signed formula \framebox{$=0.6$}$\hspace{0.3ex}P$ indicates that $P$ has truth value $0.6$, and \framebox{$\leq 0.6$}$\hspace{0.3ex}P$ indicates that $P$ has some truth value in the interval $[0,0.6]$.

\begin{figure}[!h]
	\centering
	\includegraphics[scale=1,clip=true,trim=150 470 320 120]{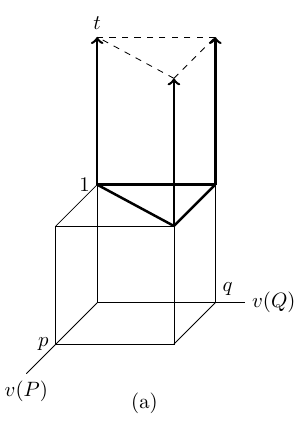} \hspace{3ex}	\includegraphics[scale=1,clip=true,trim=150 470 320 120]{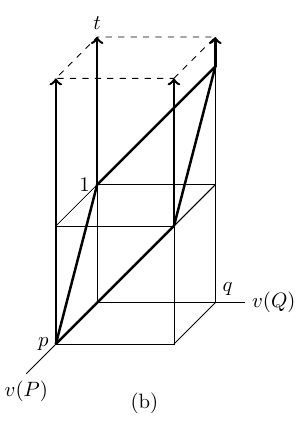}
	\caption{Two polyhedra with the same unique recession direction $(v(P),v(Q),t)=(0,0,1)$.  Heavy lines indicate the edges of the polyhedra.}
	\label{fig:MILP}
\end{figure}

To illustrate the role of MILP, assume for the moment that truth values are continuous in the interval $[0,1]$.  The resulting model is easily modified to allow only a finite number of discrete truth values.  Suppose first that we wish to infer possible truth values of $P\supset Q$ from the premises $\framebox{$\leq p$}\hspace{0.3ex} P$ and $\framebox{$\leq q$}\hspace{0.3ex}Q$. For example, we may wish to know how false $P\supset Q$ can be, given the premises.  We determine this by minimizing $t$ subject to the condition that $\framebox{$\leq t$}\hspace{0.3ex}(P\supset Q)$ can be inferred from the premises.  If $t^*$ is the minimizing $t$, then $\framebox{$\leq t$}\hspace{0.3ex}(P\supset Q)$ can be inferred for all $t\in[t^*,1]$, meaning that $P\supset Q$ cannot be more false than $t^*$.  Using \eqref{eq:multi1}, the minimization problem is
\begin{equation}
\min_{t,v(P),v(Q)} \big\{
t \; \big| \; t\geq \min\{1, 1-v(P)+v(Q)\}, \; 0\leq v(P)\leq p, \; 0\leq v(Q)\leq q
\big\}
\label{eq:multi2}
\end{equation}
To obtain an MILP model of problem \eqref{eq:multi2}, we observe first that any feasible solution $(t,v(P),v(Q))$ must satisfy the following (inclusive) disjunction:
\begin{equation}
\begin{array}{l}
t\geq 1 \\
0\leq v(P) \leq v(Q) \\
0\leq v(Q)\leq q
\end{array} 
\hspace{5ex} \mbox{or} \hspace{5ex}
\begin{array}{l}
t\geq 1-v(P)+v(Q) \\
0\leq v(Q) \leq q \\
0\leq v(P) \leq p
\end{array} \label{eq:multi3}
\end{equation}
The feasible set is therefore the union of the two polyhedra described respectively by the two systems in \eqref{eq:multi3}, and illustrated respectively by Fig.~\ref{fig:MILP}(a) and~\ref{fig:MILP}(b).  Each polyhedron has the single recession direction $(t,v(P),v(Q))=(1,0,0)$, and the necessary and sufficient condition for MILP representability is therefore satisfied.

An MILP model is now obtained as follows.  In general, it is shown in \cite{Jer89} that if the feasible set is a union of polyhedra described by $A^i\bm{x}\leq \bm{b}^i$ for $i\in I$, the MILP constraint set consists of
\begin{align*}
& A^i\bm{x}^i\leq \bm{b}^i\delta_i, \;i\in I \\
& \bm{x} = {\textstyle \sum_{i\in I} \bm{x}^i, \;\; \sum_{i\in I} \delta_i = 1} \\
& \delta_i\in \{0,1\}, \; i\in I
\end{align*}
where $\bm{x}^i$ and $\delta_i$ are new continuous and binary variables, respectively, for all $i\in I$.   An MILP model of \eqref{eq:multi2} therefore has the form
\begin{equation}
\min \left\{
t_1 + t_2 \;\left|\;
\begin{array}{l@{\hspace{3ex}}l}
\begin{array}{l}
t_1\geq\delta_1\\
0\leq v_1(P)\leq v_1(Q) \\
0\leq v_1(Q) \leq q\delta_1 
\end{array}
&
\begin{array}{l}
t_2\geq \delta_2-v_2(P)+v_2(Q) \\
0\leq v_2(Q)\leq q\delta_2 \\
0\leq v_2(P)\leq p\delta_2
\end{array} \\
\multicolumn{2}{c}{v(P)=v_1(P)+v_2(P)} \\
\multicolumn{2}{c}{v(Q)=v_1(Q)+v_2(Q)} \\
\multicolumn{2}{c}{\delta_1+\delta_2=1, \; \delta_1,\delta_2\in\{0,1\}}
\end{array}
\right.
\right\}
\label{eq:multi4}
\end{equation}
It is straightforward (if somewhat tedious) to show that \eqref{eq:multi4} simplifies to the MILP model
\begin{equation}
\min \left\{
t \;\left|\;
\begin{array}{l@{\hspace{3ex}}l}
\begin{array}{l}
t\geq\delta\\
0\leq v(P)\leq v(Q)+1-\delta \\
0\leq v(Q) \leq q
\end{array}
&
\begin{array}{l}
t\geq 1-\delta-v(P)+v(Q) \\
0\leq v(P)\leq p \\
\delta\in \{0,1\}
\end{array} 
\end{array}
\right.
\right\}
\label{eq:multi5}
\end{equation}
If we wish to recognize $m+1$ equally spaced discrete truth values in $[0,1]$ rather than continuous truth values, we can replace $v(P)$ and $v(Q)$ with $v_I(P)/m$ and $v_I(Q)/m$, respectively in \eqref{eq:multi5}, and require that $v_I(P)$ and $v_I(Q)$ be integers.  Then if $t^*$ is the optimal solution of the resulting MILP problem, we can infer $\framebox{$\leq t$}\hspace{0.3ex}(P\supset Q)$ for all $t\in \{t^*/m,(t^*\!+\!1)/m,\ldots,1\}$. 

The MILP model for more complex formulas can be constructed recursively.  For example, the formula $(P\!\supset\!Q)\supset \neg P$ has the truth function.  
\[
v\big((P\!\supset\!Q)\supset\neg P\big) = 
\min \big\{ 1, \; 1-\min\{1, \; 1\!-\!v(P)\!+\!v(Q)\} + 1\!-\!v(P)\big\}
\]
An MILP model can be written by allowing the constraints in \eqref{eq:multi5} to represent the possible truth values $t$ of the inner implication $P\supset Q$, and observing that the truth value of $\neg P$ is $1-v(P)$.  Then we can write additional constraints to represent the possible truth values $t'$ of the outer implication.  Using $\delta'$ as the binary variable in the latter, this yields the model
\begin{equation}
\min \left\{
t' \;\left|\;
\begin{array}{l@{\hspace{3ex}}l}
\begin{array}{l}
t'\geq\delta'\\
0\leq t \leq (1\!-\!v(P)) +1\!-\!\delta' \\
\end{array}
&
\begin{array}{l}
t'\geq 1\!-\!\delta'-t+(1\!-\!v(P)) \\
0 \leq t \leq 1, \;\; \delta'\in\{0,1\}
\end{array} \\
\multicolumn{2}{c}{\mbox{constraints in \eqref{eq:multi5}}}
\end{array}
\right.
\right\}
\label{eq:multi6}
\end{equation}
The valid bound $t\leq 1$ ensures that the polyhedra whose union \eqref{eq:multi6} represents have the same recession direction $(t',t,v(P),v(Q))=(1,0,0,0)$.  

As an example, suppose the upper bounds on $v(P)$ and $v(Q)$ are $(p,q)=(0.3,0.7)$.  The minimum value of $t'$ is 0.7, indicating that $(P\!\supset\! Q)\!\supset\!\neg P$ cannot be more false than 0.7.  This solution is achieved when $(t,v(P),v(Q))=(1,0.3,0.3)$ and $(\delta,\delta')=(0,0)$.  A branch-and-bound solution of this problem instance appears in Section~\ref{sec:inferencePostop}.

Additional discussion of MILP models for multivalued logic can be found in \cite{ChaHoo99,Hah94}.

\section{Statistical Inference of Logical Formulas}

As noted in the introduction, the estimation of weights in a neural network poses an optimization problem.  The same is true for the calibration of a vector support machine, which is a hyperplane that classifies by separating two (possibly overlapping) clusters of data points.  There is a third data fitting possibility that is often overlooked.  One can fit a {\em logical formula} to data in a manner analogous to classical regression, which likewise gives rise to an optimization problem.  This represents yet another instance in which logic and optimization combine to serve the purposes of AI.  Like logic-based methods in general, it has the advantage of transparency, since one can examine precisely what rules are inferred from training data.  It also offers the possibility of computing the statistical significance of an inferred formula, as is routinely done in regression analysis.  A similar approach can be used to reduce a trained neural network to a logical formula that approximates its predictions, for the sake of transparency, by fitting the formula to the output of the network for various inputs.

\subsection{Boolean Regression}

Boolean regression \cite{BorHamHoo95} infers a logical formula from Boolean data points of the form $(\bm{x},y)$, where the independent variables $\bm{x}=(x_1,\ldots,x_m)$ represent observed attributes and the dependent variable $y$ represents the observed outcome.  Following classical regression analysis, it supposes there is noise in the observations and therefore writes
\[
y = f(\bm{x})\oplus \epsilon
\]
where $f(\bm{x})$ is the true outcome, and $\epsilon$ is the observation error.  Here, $\oplus$ is a binary sum, so that $a\oplus b = (a+b)\bmod 2$.  The Boolean error term $\epsilon$ is a simple Bernoulli variable that takes the value 1 with probability $p$ and 0 with probability $1-p$.  The value of $p$ is estimated along with the regression formula, much as the error variance is estimated along with the regression coefficients in classical statistics.  An independent variable $x_i$ that is non-Boolean and takes discrete values $0,1, 2,\ldots,k$ can be accommodated by replacing it with Boolean variables $x_{i0},\ldots,x_{i\ell}$, where $x_i=x_{i0}+2x_{i1}+\cdots+2^{\ell}x_{i\ell}$.  

The dataset may contain multiple observations of $y$ for a given value of $\bm{x}$, and no observation of $y$ at all for other values of $\bm{x}$.  This is illustrated by an example taken from \cite{BorHamHoo95}, whose dataset (Table~\ref{ta:boolean1}) is designed to contain a great deal of noise.  We wish to identify a Boolean function $f(\bm{x})$ that best fits the data, where $f$ belongs to a specified class $F$ of Boolean functions, such as the class of functions expressed by a Boolean formula of a certain form.

\begin{table}[!h]
\centering
\caption{Small dataset for Boolean regression.}
\label{ta:boolean1}
\begin{tabular}{ccccc@{\hspace{4ex}}cc}
\hline \\ \ \\[-4.7ex]
\multicolumn{5}{@{\hspace{-2ex}}c}{Observations} & \multicolumn{2}{@{\hspace{-2ex}}c}{Number of} \\
\multicolumn{5}{@{\hspace{-2ex}}c}{of $\bm{x}$} & \multicolumn{2}{@{\hspace{-2ex}}c}{observations with} \\
$x_1$&$x_2$&$x_3$&$x_4$&$x_5$& $y=0$ & $y=1$ \\
\hline \\ \ \\[-4.7ex]
0 & 1 & 0 & 1 & 1 & 7 & 15 \\
1 & 1 & 0 & 0 & 1 & 9 & 3 \\
1 & 0 & 0 & 1 & 1 & 8 & 2 \\
1 & 0 & 0 & 0 & 1 & 3 & 7 \\
1 & 1 & 0 & 1 & 0 & 5 & 17 \\
0 & 0 & 1 & 1 & 1 & 9 & 2 \\
\hline
\end{tabular}
\end{table}

A classical least-squares fit is based on maximum likelihood estimation, and a similar approach can be used here.  We seek a function $f$ that maximizes the likelihood of the errors displayed in Table~\ref{ta:boolean1}.  Thus if $\bm{y}=(y_1,\ldots,y_n)$ represents the set of observations, we wish to identify a function $f\in F$ and probability $p$ that maximize the likelihood function
\begin{equation}
\mathrm{Pr}(\bm{y}|f,p) = (1-p)^{n-e(f)}p^{e(f)}
\label{eq:boolean2}
\end{equation}
where $n$ is the number of observations, and $e(f)$ is the number of errors made by function $f$ on the observed values of $\bm{x}$.  For example, if $(\bm{x}^1,\ldots,\bm{x}^6)$ are the observed values of $\bm{x}$ in Table~\ref{ta:boolean1}, and $(f(\bm{x}^1),\ldots,f(\bm{x}^6))=(0,\ldots,0)$, then $Pr(\bm{y}|f,p) = (1-p)^{41}p^{46}$.  If $(f(\bm{x}^1),\ldots,f(\bm{x}^6))=(0,0,0,0,0,1)$, then $Pr(\bm{y}|f,p) = (1-p)^{34}p^{53}$, and so forth.  

We can note right away that \eqref{eq:boolean2} is maximized with respect to $p$ when $p=e(f)/n$.   It therefore suffices to maximize the following over $f\in F$:
\begin{equation}
\mathrm{Pr}\big(\bm{y}\big|f,e(f)/n\big) = \Big(1 - \frac{e(f)}{n}\Big)^{n-e(f)} \Big(\frac{e(f)}{n}\Big)^{e(f)}
\label{eq:boolean5}
\end{equation}
Interestingly, since \eqref{eq:boolean5} is convex in $e(f)$, it is maximized either by a function $f$ that minimizes $e(f)$ or by a function that maximizes $e(f)$.  Most applications call for a fit that minimizes errors, but there are cases in which it may be appropriate to maximize errors \cite{BorHamHoo95}.  For present purposes, we assume that the data are such that a maximum likelihood fit is error minimizing.  If $\hat{f}$ is the error minimizing function, then $\hat{p}=e(\hat{f})/n$ is a (biased) estimator of $p$.

Suppose now that we wish to fit to the data of Table~\ref{ta:boolean1} a regression formula that is a logical clause with positive literals:
\begin{equation}
\beta_0 \vee \beta_1x_1 \vee\cdots\vee \beta_5x_5
\label{eq:boolean4}
\end{equation}
The binary coefficient $\beta_i=1$ (for $i\geq 1$) indicates that $x_i$ is present in the clause, and $\beta_i=0$ indicates that $x_i$ is absent.  The clause is a tautology if $\beta_0=1$, and $\beta_0$ disappears from the clause if $\beta_0=0$.  The task is to find a maximum likelihood estimate of $\beta_0,\ldots,\beta_5$, which is accomplished by maximizing \eqref{eq:boolean5} over all functions $f$ that result from some setting of $\beta_0,\ldots,\beta_5\in\{0,1\}$.  

The maximization problem can be solved by formulating it as a pseudo-Boolean optimization problem, mentioned earlier in the context of probabilistic logic.  Let $f_{\bm{\beta}}$ be the Boolean function that results from a given tuple $\bm{\beta}=(\beta_0,\ldots,\beta_5)$ of parameters in \eqref{eq:boolean4}.  We first note that if 
\begin{equation}
\beta_0=\beta_2=\beta_4=\beta_5=0
\label{eq:boolean6}
\end{equation}
then $f_{\bm\beta}(\bm{x}^1)=0$, and line~1 of Table~\ref{ta:boolean1} generates 15 errors.  If \eqref{eq:boolean6} does not hold, then $f_{\bm\beta}(\bm{x}^1)=1$, and there are 7 errors.  So, line~1 generates
\[
15\bar{\beta}_0\bar{\beta}_2\bar{\beta}_4\bar{\beta}_5
+ 7(1-\bar{\beta}_0\bar{\beta}_2\bar{\beta}_4\bar{\beta}_5) 
\]
errors, where $\bar{\beta}_i=1-\beta_i$.  Summing expressions of this sort for $\bm{x}^1,\ldots,\bm{x}^6$ and collecting terms, we obtain the pseudo-Boolean function $e(f_{\bm\beta})$ equal to
\[
41 + 8\bar{\beta}_0\bar{\beta}_2\bar{\beta}_4\bar{\beta}_5
- 6\bar{\beta}_0\bar{\beta}_1\bar{\beta}_2\bar{\beta}_5
-6\bar{\beta}_0\bar{\beta}_1\bar{\beta}_4\bar{\beta}_5
+4\bar{\beta}_0\bar{\beta}_1\bar{\beta}_5
+12\bar{\beta}_0\bar{\beta}_1\bar{\beta}_2\bar{\beta}_4
-7\bar{\beta}_0\bar{\beta}_3\bar{\beta}_4\bar{\beta}_5
\]
The minimum error solution is $\bm{\beta}=(0,0,1,0,0,0)$, with $e(f_{\bm\beta})=32$ errors, which means the estimated error probability is $\hat{p}=e(f_{\bm\beta})/n=32/87\approx 0.368$.  This solution corresponds to the simple logical formula $x_2$, which ignores all but one of the independent variables.  To obtain a more satisfactory formula, one might allow negative as well as positive literals:
\begin{equation}
\beta_0 \vee \beta_1x_1 \vee\cdots\vee \beta_5x_5
\vee \gamma_1(\neg x_1) \vee\cdots\vee \gamma_5(\neg x_5)
\label{eq:boolean25}
\end{equation}
The best-fit formula is $x_2\vee\neg x_4$, which results in $e(f_{\bm\beta,\bm\gamma})=28$ errors and $\hat{p}\approx 0.322$. 

In general, the regression formula can be any expression of the form
\begin{equation}
f(\bm{x}_{\beta})= \bigvee_{i\in I} \beta_iP_i(\bm{x})
\label{eq:boolean10}
\end{equation}
where each $P_i(\bm{x})$ is any desired logical proposition that contains atomic propositions in $\bm{x}=(x_1,\ldots,x_m)$.  For example, each $P_i(\bm{x})$ could be a conjunction of literals (atomic propositions or their negations); in fact, any propositional formula can be written as \eqref{eq:boolean10} with $P_i(\bm{x})$s in this form.  If $\bm{x}^1,\ldots,\bm{x}^n$ are the observations of independent variables, let $y_0(\bm{x}^j)$ be the number of observations of $f(\bm{x}^j)$ for which $y=0$, and $y_1(\bm{x}^j)$ the number for which $y=1$. The pseudo-Boolean function to be minimized is
\[
e(f_{\bm{\beta}}) = \sum_{j=1}^n y_0({\bm x}^j) + 
\sum_{j=1}^n \big(y_1(\bm{x}^j)-y_0(\bm{x}^j)\big) \hspace{-2ex}\prod_{i|P_i(\bm{x}^j)=1}\hspace{-2.5ex}\bar{\beta}_i
\]

Solution methods for pseudo-Boolean optimization are extensively discussed in \cite{BorHam02}.  Recent developments in solution methods include \cite{ElfNor18,ElfNor20,SmiBerJar22}, and schemes for integrating MILP techniques are described in \cite{DevGleNor21,DevGocDemNorStu21}.

\subsection{Statistical Significance}

Boolean regression can provide measures of statistical significance, including confidence levels and the significance of regression, the latter of which enables stepwise regression.  A potential barrier to calculating significance is that the distribution of estimated regression parameters depends on the distribution of observational error, which is unknown {\em a priori}.  Classical regression deals with this problem by identifying a ``pivot'' statistic whose distribution depends only on the number of data points (or more precisely, the degrees of freedom).  For example, the $t$-statistic allows one to compute confidence intervals based on the pre-computed distribution of the \mbox{$t$-statistic}.  Apparently, no pivot statistic has been identified for Boolean regression, but one can nonetheless obtain useful significance indicators from a Bayesian model. We follow here the analysis of \cite{BorHamHoo95}.

A Bayesian significance model identifies the confidence level of a deduced regression formula $f$ with the posterior probability that $f$ is the true formula, given the observed data and prior probabilities.  
Thus, given observations $\bm{y}=(y_1,\ldots,y_n)$, the confidence level of $f$ is Pr$(f|\bm{y})$.  Using Bayes' rule, this is
\begin{equation}
\mathrm{Pr}(f|\bm{y}) = \frac{\mathrm{Pr}(\bm{y}|f)\mathrm{Pr}(f)} {\mathrm{Pr}(\bm{y})}
\label{eq:boolean20}
\end{equation}
where Pr$(f)$ is the prior probability that function $f$ is the correct one.  The conditional probability Pr$(\bm{y}|f)$ is obtained by integrating over possible error probabilities $p$:
\begin{equation}
\mathrm{Pr}(\bm{y}|f) = \int_0^1 \mathrm{Pr}(\bm{y}|f,p)\pi(p)dp
\label{eq:boolean21}
\end{equation}
where $\pi(p)$ is a prior probability density function for $p$, and Pr$(\bm{y}|f,p)$ is given by \eqref{eq:boolean2}.  Also Pr$(\bm{y})$ is 
\[
\mathrm{Pr}(\bm{y}) = \sum_{f'\in F} \mathrm{Pr}(\bm{y}|f')\mathrm{Pr}(f')
\]
Since nothing is known in advance about which formula could be correct, we assume Pr$(f')$ is the same for all $f'\in F$ and can therefore ignore this term.  As for the probability $p$ of error, we can suppose it is no greater than some value $\rho\leq \half$, since even a random guess is right half the time.  Since we know nothing beyond this, we assume $p$ is uniformly distributed over $[0,\rho]$, with $\pi(p)=1/\rho$ in this range.  Thus, using \eqref{eq:boolean2}, \eqref{eq:boolean21} becomes
\[
\mathrm{Pr}(Y|f)=\frac{1}{\rho}\int_0^\rho (1-p)^{n-e(f)}p^{e(f)} dp
\]
Evaulating the integral, we obtain
\begin{equation}
Pr(\bm{y}|f) = \frac{\rho^{e(f)}}{n+1} \sum_{i=0}^{n-e(f)} 
\hspace{-0.5ex} \Big( \begin{array}{@{}c@{}}n-e(f) \\ i \end{array} \Big)
\Big( \begin{array}{@{}c@{}} n \\ i \end{array} \Big)^{\!-1}
\hspace{-1.5ex} (1-\rho)^{n-e(f)-i}
\end{equation}

It is reported in \cite{BorHamHoo95} that, using this analysis, the maximum likelihood formula $x_2$ can be inferred from the data in Table~\ref{ta:boolean1} with confidence level 0.282 when $\rho=\half$.   This says one cannot have much confidence that the formula is exactly right.  One might calculate a confidence sphere that is analogous to an confidence interval in classical regression.  A confidence sphere could consist of all binary vectors $\bm{\beta}$ within a Hamming distance of $d$ from the inferred vector $\bm{\beta}$ of coefficients.  In the example, the confidence level increases to 0.433 when $d=1$ and $0.572$ when $d=2$, still rather low due to the large amount of noise in the dataset.

The significance of regression, however, is more encouraging.  We can compute it by considering the hypotheses:
\begin{align*}
& H_0: \; \beta_i=0\;\mbox{for}\; i=1,\ldots, 5 \; \mbox{(null hypothesis)}\\
& H_1: \; \beta_i=1\;\mbox{for at least one}\; i\in\{1,\ldots,5\}
\end{align*}
The best fit under the null hypothesis is the function $f_0$ defined by $\beta_0=1$, with Pr$(f_0|\bm{y})=0.0104$.  The best fit under $H_1$ is again the function $f$ corresponding to $\bm{\beta}=(0,0,1,0,0,0)$ with Pr$(f|\bm{y})=0.282$.  The significance of regression is
\[
\frac{\mathrm{Pr}(f_0|\bm{y})} {\mathrm{Pr}(f_0|\bm{y})+\mathrm{Pr}(f|\bm{y})} = 0.036
\]
This means that it is very likely (96.4\% probability) that an expression of the form $\beta_1x_1\vee\cdots\vee\beta_5x_5$ captures a relationship that is not purely the result of chance.   

We can also perform stepwise regression by adding negative literals to the formula as in \eqref{eq:boolean25}.  This significantly improves the fit if we can reject $H_0$:
\begin{align*}
& H_0: \; \beta_i=0\;\mbox{for} \;i=6,\ldots,10 \; \mbox{(i.e., the negative literals do not improve the fit)} \\
& H_1: \; \beta=1 \; \mbox{for at least one}\;i\in \{6,\ldots,10\}
\end{align*}
We can reject $H_0$ with probability $1-0.073=0.927$, which means that the improvement in fit is significant at the 10\% level but not at the 5\% level.  In addition, the confidence level of the expanded regression formula $x_2\vee\neg x_4$ is only 0.11, because it is competing with a much larger collection of admissible formulas.  

Related research on learning Boolean functions appears in \cite{Ant10,CheBra17,MukPelNevKuoZiySpeGraSpe09,MukSpe07,SanTriCheLia02,SloSzoTur10}.  An important class of Boolean functions are monotone, meaning that $\bm{x}'\geq \bm{x}$ implies $f(\bm{x}')\geq f(\bm{x})$, where $\bm{x}'\geq \bm{x}$ when $x'_j\geq x_j$ for each $j$.  For example, a monotone Boolean function implies that observing additional indicators of a particular disease cannot reverse a prediction that the disease is present.  The problem of learning monotone Boolean functions from noisy and incomplete data is addressed in \cite{AmaMar06,BluBurLan98,BorHamHoo94}.

\section{Inference as Projection}  \label{sec:projection}

Optimization and logical inference are fundamentally related because they are both special cases of projection \cite{Hoo16}.  The {\em projection} of a set $S$ of tuples $\bm{x}=(x_1,\ldots, x_n)$ onto variables $\bar{\bm{x}}=(x_1,\ldots,x_k)$ is 
\[
S|_{\bar{\bm{x}}}=\big\{(x_1,\ldots,x_k)\;\big|\;(x_1,\ldots,x_n)\in S\big\}
\]
Optimization is a special case of projection because any optimization problem can be written $\min/\max \{z\;|\; (z,\bm{x})\in S\}$ and solved by computing the projection $S|_z$.   The minimum and maximum values of $z$, if they exist, are then easily recognized as the smallest and largest elements of $S|_z$.  Inference can be seen as a special case of projection if we let $S(F)$ denote the set of assignments to $\bm{x}$ that satisfy a set $F$ of logical formulas.  Then a formula $f$ containing variables $\bar{\bm{x}}$ can be inferred from $F$ if and only if all the assignments in the projection $S(F)|_{\bar{\bm{x}}}$ satisfy $f$.

This linkage of optimization and inference through projection provides a novel opportunity to apply optimization to inference.  The projection of $S(F)$ onto $\bar{\bm{x}}$ can be obtained with the assistance of a binary decision diagram (BDD), which compactly and transparently displays satisfying solutions of $F$.  The projection can be computed using a generalization of Benders decomposition, a well-known optimization method that, in effect, (partially) computes the projection of the feasible set onto a subset of variables.   The first subsection below illustrates how a BDD represents the satisfying set $S(F)$, and how it relates to projection.  The second subsection shows how Benders decomposition can accelerate the computation of the projection, thereby deducing all formulas that can be inferred from $F$.  

This methodology can be useful to AI because it allows one to infer everything that can be deduced from a knowledge base regarding a specific topic, rather than check which individual formulas can be deduced.  For example, one may be interested in whether substances A, B, and C can interact to cause allergic reactions D and E.  Rather than check whether one can deduce that A and B interact to cause D (i.e. $(A\wedge B)\supset D$), whether one can deduce that B and C cause reaction E, and so forth, one can project the knowledge base onto the five variables A, B, C, D, and E.  Since one is projecting onto a handful of variables, the projection can be computed fairly rapidly.  The rather small number of assignments in the projection (at most $2^5$) is easily converted to a few logical formulas that capture everything that is known about allergic reactions to A, B and C.

\subsection{Projection in a Binary Decision Diagram} \label{sec:projectionBDD}

Binary decision diagrams were introduced in the 1970s and have been used for logic circuit verification, product configuration, and more recently, discrete optimization \cite{Ake78,BerCirHoeHoo16a,BolSauSieWeg2010,Bry92,PopBalOst23}.  
We are interested in their potential for computing projections.  
A BDD is a graphical representation of a Boolean function $f(\bm{x})$.  Any set $F$ of propositional formulas defines a Boolean function $f(\bm{x})$ that takes the value 1 when $\bm{x}$ satisfies $F$ and 0 otherwise, where $\bm{x}=(x_1,\ldots,x_n)$ are the variables in $F$.  Thus, any set $F$ of formulas is represented by a suitable BDD.

\begin{figure}[!h]
	\centering
	\includegraphics[scale=1,clip=true,trim=10 10 10 10]{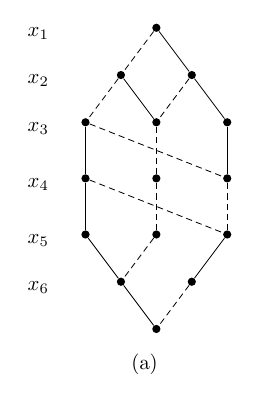} \hspace{7ex}
	\includegraphics[scale=1,clip=true,trim=10 10 10 10]{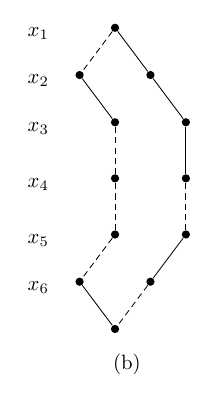} \hspace{7ex}
	\includegraphics[scale=1,clip=true,trim=10 10 10 10]{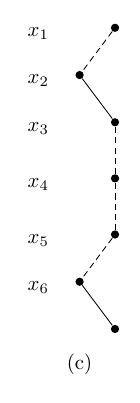} 
	\caption{(a) Reduced binary decision diagram for a small set of logical propositions.  (b) Reduced BDD after fixing $x_2=1$.  (c) Reduced BDD after fixing $(x_2,x_4,x_6)=(1,0,1)$.}
	\label{fig:proj1}
\end{figure}

Consider, for example, the set $F$ of formulas 
\begin{equation}
\begin{array}{l@{\hspace{4ex}}l}
x_1x_2 \supset (x_3\bar{x}_4x_5\bar{x}_6) & (f_1) \\
\bar{x}_1\bar{x}_2 \supset (\bar{x}_4x_5\bar{x}_6 \vee x_3x_4x_5x_6) & (f_2)\\
(x_1\bar{x}_2 \vee \bar{x}_1x_2)\supset
\bar{x}_3\bar{x}_4\bar{x}_5x_6 & (f_3) 
\end{array} \label{eq:proj10}
\end{equation}
where, for readability, $x_ix_j$ means $x_i\wedge x_j$ and $\bar{x}_i$ means $\neg x_i$.  The set $F$ is represented by the {\em ordered BDD}\footnote{Often abbreviated OBDD.  We use the abbreviation BDD since only ordered BDDs are of interest here.}  in Fig.~\ref{fig:proj1}(a).  The nodes of the BDD are arranged in layers that, except for the terminal node at the bottom, correspond to variables $x_1,\ldots,x_6$.  A dashed arc leaving a node in layer $i$ toward a layer below it represents setting $x_i=0$, and a solid arc represets $x_i=1$.  Each path from top to bottom therefore represents an assignment to $\bm{x}$.  The BDD is constructed so that the top-to-bottom paths correspond exactly to assignments that satisfy the formulas in $F$.\footnote{Classically, BDDs also contain paths to a second terminal node that represent non-satisfying assignments, but these can be omitted for present purposes.}  The BDD in Fig.~\ref{fig:proj1}(a) is {\em reduced} in the sense that no smaller BDD that uses the same variable ordering represents \eqref{eq:proj10}.  In fact, any given Boolean function is uniquely represented by a reduced BDD with a specified variable ordering \cite{Bry92}.  Methods for constructing reduced BDDs are described in \cite{Weg00,Weg04}.

Let's suppose that we wish to know what can be deduced from \eqref{eq:proj10} regarding the variables $x_2$, $x_4$, and $x_6$.  To deduce all formulas containing $x_2, x_4, x_6$ that are implied by \eqref{eq:proj10}, we can derive the BDD $B'$ that results from $B$ by projecting out variables $x_1,x_3,x_5$.  Then the top-to-bottom paths in $B'$ correspond to the assignments in the projection onto $(x_2,x_4,x_6)$.  The deducible formulas are then easily obtained from this projection, as we will illustrate shortly.

The classical method for projecting out a variable $x_i$ proceeds by first obtaining the BDDs $B_1$ and $B_0$ that result from setting $x_i$ to 1 and 0, respectively, and then computing $B'$ by taking the disjunction of $B_1$ and $B_0$.  However, the disjunction operation on BDDs tends to be quite expensive, and it must be performed recursively for each variable to be projected out.  

Since we wish to project \eqref{eq:proj10} onto a small number of variables, it is more efficient to check directly which assignments to $(x_2,x_4,x_6)$ are represented by top-to-bottom paths in $B$. Suppose, for example, we first check $(x_2,x_4,x_6)=(1,1,1)$.  We might begin by setting $x_2=1$ and removing the dashed arcs corresponding to $x_2=0$ in Fig.~\ref{fig:proj1}(a).  This permits the removal of several arcs and nodes because they no longer lie on a top-to-bottom path, resulting in the smaller BDD of Fig.~\ref{fig:proj1}(b).  Now if we try setting $x_4=1$ and remove all dashed arcs corresponding to $x_4=0$, this causes all remaining arcs to be removed.  Since this already eliminates all top-to-bottom paths, $(x_2,x_4,x_6)$ can be neither $(1,1,0)$ nor $(1,1,1)$.  Now if we try $(x_2,x_4,x_6)=(1,0,1)$, the BDD reduces to that in Fig.~\ref{fig:proj1}(c).  Since there is a top-to-bottom path (in this case, a single such path), we conclude that $(1,0,1)$ is part of the projection.  Continuing in this fashion, we find that the projection onto $(x_2,x_4,x_6)$ is
\begin{equation}
\big\{(1,0,1), (1,0,0), (0,1,1), (0,0,1), (0,0,0)\big\}
\label{eq:proj25}
\end{equation}
It is readily seen that these are the satisfying assignments of the formulas $x_2\supset \neg x_4$ and $x_4\supset x_6$.
These formulas therefore represent everything that can be deduced from \eqref{eq:proj10} regarding variables $x_2,x_4,x_6$.  

A possible barrier to the use of reduced BDDs is their ability to grow exponentially with the number of variables.  However, the size of the BDD depends on other factors as well.  For example, the knowledge base may decouple into sections that deal with different topics and therefore have no variables in common.  In such cases, a BDD can be generated for each section independently of the others, resulting in much smaller diagrams.   A generalization of this approach exploits the structure of the dependency graph of the knowledge base.  The dependency graph contains a vertex for every formula, and two vertices are connected by an edge when the corresponding formulas have at least one variable in common.  The complexity of logical inference is at worst exponential in the treewidth of the dependency graph, which tends to be small when the formulas are loosely coupled.  In such cases, one can build a nonserial BDD as described in \cite{Hoo13}, which can be markedly smaller than a conventional serial BDD and is quite suitable for computing projections.  In addition, the size of a serial or nonserial BDD can be very sensitive to the ordering of the variables, and it may be possible to find an ordering that results in a much smaller diagram \cite{Weg00,Weg04}.  Finally, given that AI's large language models are based on neural networks that may contain upwards of a trillion parameters, a very large BDD may be acceptable in AI applications.

\subsection{Projection with Benders Decomposition} \label{sec:projection.Benders}

The previous section showed how to compute the projection of \eqref{eq:proj10} onto $(x_2,x_4,x_6)$ by employing a decision diagram to check every possible assignment to these variables for consistency with \eqref{eq:proj10}.  It is generally possible, however, to accelerate this process significantly by applying Benders decomposition.  This popular optimization technique was originally applied to mixed integer programming \cite{Ben62}, but we use a generalization known as logic-based Benders decomposition (LBBD) \cite{Hoo00,Hoo24,HooOtt03}.  

LBBD is applied to a general optimization problem of the form 
\[
\max\big\{ 
f(\bm{x},\bm{y}) \; \big| \; (\bm{x},\bm{y})\in S, \; \bm{x}\in D_{\bm{x}}, \; \bm{y}\in D_{\bm{y}}
\big\}
\]
where $D_{\bm{x}}$ and $D_{\bm{y}}$ are the domains of $\bm{x}$ and $\bm{y}$, respectively (e.g., tuples of integers or reals).  For our purposes, we can restrict attention to problems in which the objective function depends only on $\bm{x}$:
\begin{equation}
\max\big\{ 
f(\bm{x}) \; \big| \; (\bm{x},\bm{y})\in S, \; \bm{x}\in D_{\bm{x}}, \; \bm{y}\in D_{\bm{y}}
\big\}
\label{eq:proj20}
\end{equation}
The problem is decomposed into a master problem
\[
\max\big\{f(\bm{x})\;\big|\; \mbox{Benders cuts}, \; \bm{x}\in D_{\bm{x}}\big\}
\]
and a feasibility subproblem that asks whether $(\bar{\bm{x}},\bm{y})\in S$ for some $\bm{y}\in D_{\bm{y}}$,
where $\bar{\bm{x}}$ is the most recent solution of the master problem.  If the subproblem has a feasible solution $\bar{\bm{y}}$, the original problem \eqref{eq:proj20} has optimal solution $(\bar{\bm{x}},\bar{\bm{y}})$.  If the subproblem is infeasible, the proof of infeasibility is analyzed to obtain a valid constraint (Benders cut) $\bm{x}\in S_{\bar{\bm{x}}}$ that is violated by $\bm{x}=\bar{\bm{x}}$ as well as by other values of $\bm{x}$ that the same proof shows to be infeasible.  The Benders cut is added to the constraint set of the master problem, which is re-solved to obtain the next $\bar{\bm{x}}$.  The process continues until (a) the subproblem is feasible, or (b) the master problem is infeasible (in which case the original problem infeasible).  

Of greatest interest to us is the fact that Benders cuts generated during this process at least partially define the projection of the feasible set of \eqref{eq:proj20} onto $\bm{x}$.  A complete description of the projection can be obtained simply by adding the constraint $\bm{x}\neq \bar{\bm{x}}$ to the master problem (and continuing to generate cuts) each time an optimal solution $(\bar{\bm{x}},\bar{\bm{y}})$ is found. 
When the master problem finally becomes infeasible, the accumulated Benders cuts completely describe the projection onto $\bm{x}$.  

To project \eqref{eq:proj10} onto $(x_2,x_4,x_6)$, we solve the problem 
\begin{equation}
\max\big\{x_2+x_4+x_6\;\big|\; \eqref{eq:proj10}\big\}
\label{eq:proj22}
\end{equation}
by LBBD as follows.\footnote{We could just as well minimize, or replace any $x_i$ by $-x_i$.}  We let $\bm{x}=(x_2,x_4,x_6)$ and $\bm{y}=(x_1,x_3,x_5)$, so that the master problem is the integer programming problem
\begin{equation}
\max \big\{x_2+x_4+x_6\;\big|\; \mbox{Benders cuts}, \; x_2,x_4,x_6\in\{0,1\} \big\}
\label{eq:proj21}
\end{equation}
Initially there are no Benders cuts, and the solution is $(x_2,x_4,x_6)=(1,1,1)$.  This defines a subproblem that seeks a feasible solution of \eqref{eq:proj22} in which $(x_2,x_4,x_6)=(1,1,1)$.  That is, we seek top-to-bottom path in the BDD of Fig.~\ref{fig:proj1}(a) in which arcs corresponding to $(x_2,x_4,x_6)=(0,0,0)$ are removed.  We solved this subproblem in the previous section by first fixing $x_1=1$ and generating the BDD of Fig.~\ref{fig:proj1}(b).  At this point we noted that $(x_2,x_4)=(1,1)$ is already infeasible.  This allows us to create a Benders cut (specifically, a {\em nogood constraint}) that excludes solutions in which $(x_2,x_4)=(1,1)$.   We write the cut as $(1-x_2)+(1-x_4)\geq 1$ and add it to the master problem \eqref{eq:proj21}.  Re-solving the master problem, we obtain $(x_2,x_4,x_6)=(1,0,1)$, which defines a feasible subproblem and therefore solves \eqref{eq:proj22}.  To continue generating Benders cuts, we add the constraint $(x_2,x_4,x_6)\neq (1,0,1)$ to the master problem by writing $(1-x_2)+x_4+(1-x_6)\geq 1$.  After generating the optimal solutions \eqref{eq:proj25} in this fashion, the master problem becomes infeasible, and we have the projection onto $(x_2,x_4,x_6)$.  In larger instances, the Benders cuts speed up this process by eliminating many infeasible assignments to the master problem variables before they are submitted to the subproblem.

\section{Transparency through Postoptimality Analysis}

Optimization-based inference can contribute significantly to transparency in AI applications by means of postoptimality analysis.  This type of analysis can identify constraints that are essential to proving optimality, and in many cases, whether or how much the optimal solution would change if the constraints were modified a given amount.  When applied to inference, postoptimality analysis can identify premises in the knowledge base that play a role in deriving inferred propositions, and perhaps determine whether and how much the inferred propositions would change if the premises were modified.  

We begin with a brief discussion of postoptimality analysis in linear programming, which we applied in earlier sections to probabilistic logic and various belief logics.  We then describe two methods of postoptimality analysis for mixed integer/linear programming (MILP), which we applied to nonmonotonic logic, multivalued logic, Boolean regression, and inference via projection.  One is an inference-based method that uses dual multipliers obtained in the branch-and-bound search tree.  The other is based on a BDD that encodes near-optimal solutions of the problem.

\subsection{Postoptimality Analysis in Linear Programming}

Postoptimality analysis in linear programming (LP) consists primarily of elementary techniques that have been implemented in LP solvers for decades. An LP problem has the form
\begin{equation}
\min \big\{ \bm{c}^{\intercal}\bm{x} \;\big|\; A\bm{x}\geq \bm{b}, \; \bm{x}\geq \bm{0} \big\}
\label{eq:post10}
\end{equation}   
The corresponding {\em dual} problem is
\[
\max \big\{\bm{u}^{\intercal}\bm{b} \;\big| \; \bm{u}^{\intercal}A \leq \bm{c}^{\intercal}, \; \bm{u}\geq \bm{0} \big\}
\]
The dual variables $\bm{u}=(u_1,\ldots,u_m)$ correspond to the $m$ constraints $A\bm{x}\geq \bm{b}$ of the original problem \eqref{eq:post10}.  Under weak conditions, an LP problem has the same optimal value $z^*$ as its dual, and the dual solution is obtained for free as a byproduct of solving the original problem.  

The optimal dual solution $\bm{u}$ is a key to postoptimality analysis.  The dual multiplier $u_i$ associated with constraint $i$ of $A\bm{x}\geq \bm{b}$ tells us that if the right-hand side $b_i$ of constraint $i$ is changed to $b_i+\Delta b_i$ (where $\Delta b_i$ can be negative), the resulting optimal value is at least $z^*+u_i\Delta b_i$.  Also, one can easily compute a range of perturbations $\Delta b_i$ within which the new optimal value is {\em exactly} $z^*+u_i\Delta b_i$.  A consequence of this is that only constraints with positive dual multipliers $u_i$ serve as premises in the proof of optimality, and all other constraints can be dropped without changing the optimal solution.  In fact, it is not hard to state conditions under which this set of premises is irreducible, meaning that they are all essential to the proof.  This and related matters are studied in \cite{Chi96,Chi97,Chi01,ChiDra91}.  

As an example, consider the probabilistic logic problem instance discussed in Section~\ref{sec:probBasic}, in which probablities 0.9, 0.8, and 0.7 are assigned to $x_1$, $x_1\supset x_2$, and $x_2\supset x_3$, respectively.  The inferred probability range for $x_3$ is $[0.5,0.7]$.  The lower limit 0.5 is obtained by minimizing Pr$(x_3)$, and the resulting dual multipliers for the three probability assignments are all 1.25.  So, reducing the assigned probability 0.9 to 0.8, for example, results in new a lower limit of at least $0.5-(1.25)(0.1)=0.375$.  In fact, the lower limit is exactly 0.375, because this calculation is exact for any perturbation $\Delta b_1\in [-0.4,0]$.  As it happens any positive perturbation results in an infeasible problem, and similarly for the other two assigned probabilities.\footnote{The calculated value remains a valid lower bound on the optimal value, because the optimal value of an infeasible minimization problem is $\infty$.}  The upper limit 0.7, which results from maximizing Pr$(x_3)$, yields dual multipliers 0, 0, and 1 for the three assigned probabilities.  This means that only the proposition $x_2\supset x_3$ plays a role in inferring the upper limit on Pr$(x_3)$, and the other two probability assignments can be dropped without affecting it.  

\subsection{Inference-Based MILP Postoptimality Analysis} \label{sec:inferencePostop}

An inference-based method \cite{DawHoo00} provides comprehensive postoptimality analysis for MILP when the problem has moderate size, and a limited analysis for larger instances.  We will suppose the integer-valued variables in the problem are binary, in which case it has the form
\begin{equation}
\min \Big\{ z=\bm{c}^{\intercal}\bm{x} \;\big| \; A\bm{x}\geq \bm{b}, \; \bm{x}\geq\bm{0}, \; x_i\in \{0,1\}\;\mbox{for}\; i\in I \big\}
\label{eq:post30}
\end{equation}
It is  convenient to suppose that $A\bm{x}\geq\bm{b}$ contains constraints $x_i\leq 1$ for $i\in I$.  Such problems are normally solved by a branch-and-bound method that builds a search tree by branching on binary variables.  For simplicity of exposition, we assume no cutting planes are added to the constraint set during the search.  Let $z^*$ be the optimal value of $z$ obtained for \eqref{eq:post30}.

Inference-based postoptimality analysis is based on dual solutions obtained at nodes of the search tree.  An LP relaxation of \eqref{eq:post30} is solved at each node of the tree, namely
\begin{equation}
\min \Big\{ z_{\mathrm{LP}}=\bm{c}^{\intercal}\bm{x} \;\big| \; A\bm{x}\geq \bm{b}, \; \bm{x}\geq\bm{0}, \;D\bm{x}\geq \bm{d} \big\}
\label{eq:post31}
\end{equation}
where $D\bm{x}\geq \bm{d}$ consists of branching constraints of the form $x_i\leq 0$ (when branching to $x_i=0$) or $x_i\geq 1$ (when branching to $x_i=1$) that are added along the path from the root of the tree to the current node.  If one or more of the 0--1 variables has a fractional value in the LP solution, the search creates two branches by setting $x_i=0$ and $x_i=1$ for one such variable $x_i$.  

The LP dual of \eqref{eq:post31} is
\[
\max\big\{ \bm{u}^{\intercal}\bm{b} + \bm{v}^{\intercal}\bm{d} \; 
\big|\; \bm{u}^{\intercal}A + \bm{v}^{\intercal}D \leq \bm{c}^{\intercal}, \;\bm{u}\geq\bm{0}, \; \bm{v}\geq\bm{0}\big\}
\]
Let $\bm{x}_{B}$ be the variables fixed by $B\bm{x}\geq\bm{b}$, and $\bar{\bm{x}}$ the values to which they are fixed.  Then at each leaf node of the tree, one of three cases obtains:
\begin{description}
	\item (a) Problem \eqref{eq:post31} is infeasible.  Then there is a dual feasible  extreme ray $(\bar{\bm{u}},\bar{\bm{v}})$ for which $\bm{x}_{B}=\bar{\bm{x}}_B$ violates the surrogate inequality $\bar{\bm{u}}^{\intercal}A\bm{x}\geq\bar{\bm{u}}^{\intercal}\bm{b}$.  \smallskip
	\item (b) Problem (50) has an optimal solution $\bar{\bm{x}}$ with value $\bar{z}_{\mathrm{LP}}$ in which $\bar{x}_j\in\{0,1\}$ for $j\in J$, which means that $\bar{\bm{x}}$ is feasible in \eqref{eq:post30} and therefore a candidate solution. Then if $(\bar{\bm{u}},\bar{\bm{v}})$ is an optimal dual solution, $\bm{x}_{B}=\bar{\bm{x}}_B$ violates the surrogate inequality $\bar{\bm{u}}^{\intercal}A\bm{x} -  \bm{c}^{\intercal}\bm{x} >\bar{\bm{u}}^{\intercal}\bm{b}-\bar{z}_{\mathrm{LP}}$.\smallskip
	\item (c) Problem \eqref{eq:post31} has an optimal solution with value $\bar{z}_{\mathrm{LP}}$ that is no smaller than the value $\bar{z}_{\min}$ of the best candidate solution found so far.  Then if $(\bar{\bm{u}},\bar{\bm{v}})$ is an optimal dual solution, $\bm{x}_{B}=\bar{\bm{x}}_B$ violates the surrogate inequality $\bar{\bm{u}}^{\intercal}A\bm{x} -  \bm{c}^{\intercal}\bm{x} >\bar{\bm{u}}^{\intercal}\bm{b}-\bar{z}_{\min}$.
\end{description}
To conduct postoptimality analysis, we first observe that if a constraint $i$ has a vanishing dual multiplier $\bar{u}_i$ at every leaf node, then the constraint plays no role in the proof of optimality (or infeasibility, but for simplicity we suppose that the problem is feasible).  Constraint~$i$ can therefore be dropped without changing the optimal solution. 

Beyond this, we can study which perturbations in the problem data allow one to prove an optimal value of at least $z^*$.  We first observe that setting $\bm{x}_{\bm{B}}=\bar{\bm{x}}_{\bm{B}}$ at a leaf node falsifies a logical clause that negates these settings.  For example, if a leaf node is reached by setting $(x_1,x_2,x_3)=(1,0,1)$, the clause $\neg x_1 \vee x_2 \vee \neg x_3$ is falsified.  We will call this the branching clause at the node.  Since the branching search is exhaustive, the set of branching clauses for all the leaf nodes is unsatisfiable.  We also observe that since $\bm{x}_{\bm{B}}=\bar{\bm{x}}_{\bm{B}}$ violates the surrogate inequality at any given leaf node, the surrogate inequality must imply a logical clause that implies the branching clause at that node.  We will call this the surrogate clause.  To continue the example, if the surrogate inequality at the leaf node is $-4x_1+3x_2-2x_3\geq -3$, it implies the surrogate clause $\neg x_1 \vee x_2$, which implies the branching clause $\neg x_1\vee x_2\vee \neg x_3$.  We conclude that after a perturbation, the original search tree remains exhaustive and proves an optimal value of at least $z^*$ if the perturbed surrogate inequality still implies the original surrogate clause at each node.  The actual optimal value of the perturbed problem may, of course, be larger than $z^*$.   

\begin{figure}[!h]
	\centering
	\includegraphics[scale=0.95,clip=true,trim=10 10 10 10]{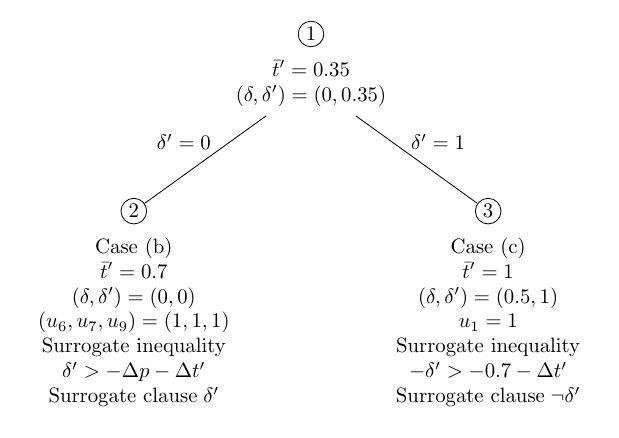} \hspace{2ex}
	\caption{Branch-and-bound search tree for the multi-valued logic problem \eqref{eq:multi6}.  Only nonzero dual multipliers $u_i$ are indicated.  The surrogate inequalities reflect a perturbation of the problem data.}
	\label{fig:postOpt}
\end{figure}

This analysis leads to a system of inequalities whose satisfaction by a perturbation ensures that the optimal value will not fall below $z^*$.
For present purposes, it suffices to illustrate how this works in the many-valued logic example presented in Section~\ref{sec:manyvalued}.  Recall that the objective is to find the smallest truth value $t'$ for which we can infer $\framebox{$\geq t'$}\,((P\supset Q)\supset \neg P)$, given \framebox{$\leq p$}$\,P$ and \framebox{$\leq q$}$\,Q$.  For the sake of illiustration, we suppose $(p,q)=(0.3,0.7)$.  The smallest $t'$ is found by minimizing $t'$ subject to the MILP constraint set in \eqref{eq:multi6}, which is reproduced below in a consistent format (omitting nonnegativity constraints on the variables): 
\begin{equation}
\begin{array}{ll@{\hspace{5ex}}ll}
\begin{array}{l}
1. \;\; t'-\delta' \geq 0 \\
2. \;\; -t +v(P)-\delta'\geq -2 \\
3. \;\; t-\delta\geq 0\\
4. \;\; -v(P)+v(Q)-\delta \geq -1\\
5. \;\; -v(Q) \geq -0.7
\end{array}
&
\begin{array}{l}
(u_1) \\ (u_2) \\ (u_3) \\ (u_4) \\ (u_5) 
\end{array}
&
\begin{array}{l}
6. \;\; t'+\delta'+t+v(P)\geq 2 \\
7. \;\; -t\geq -1 \\
8. \;\; t + v(P) -v(Q) +\delta  \geq 1 \\
9. \;\; -v(P)\geq -0.3 \\
\;\;\;\;\; \delta,\delta'\in \{0,1\}
\end{array} 
&
\begin{array}{l}
(u_6) \\ (u_7) \\ (u_8) \\ (u_9) \\ \
\end{array}
\end{array}
\label{eq:post40}
\end{equation}
The dual multipliers $u_i$ are indicated to the right of the constraints.  The search tree for this problem consists of three nodes, as shown in Fig.~\ref{fig:postOpt}.  Since $\delta'$ is a fraction 0.35 in the solution of the LP relaxation at the root node, the search branches on $\delta'$ to create nodes~2 and~3.  The LP relaxation at node~2 has an integral solution with $(\delta,\delta')=(0,0)$ and optimal value $\bar{t}'_{\mathrm{LP}}=0.7$, so that case (b) applies.  The solution at node~3 is nonintegral, but since the optimal value $\bar{t}'_{\mathrm{LP}}=1$ is worse than the best previous solution value 0.7, case (c) applies, and there is no need for further branching.  The minimum value of $t'$ is therefore 0.7, and so $(P\supset Q)\supset \neg P$ has a truth value no less than 0.7.

We are particularly interested in the sensitivity of the solution to changes in truth value bounds $(p,q)=(0.3,0.7)$, which are encoded in constraints~5 and~9.  Figure~\ref{fig:postOpt} indicates the nonzero dual multipliers at each leaf node, and we note that the multiplier $u_5$ corresponding to $v(Q)\leq 0.7$ is zero at both leaf nodes.  This means that the lower bound of 0.7 on the truth value of $(P\supset Q)\supset \neg P$ remains valid even if the bound on $v(Q)$ is dropped.\footnote{This does not mean that the bound on $v(Q)$ cannot affect the actual minimum value of $t'$.  For example, setting $v(Q)=0.1$ results in $\bar{t}'=0.9$.  Yet $\bar{t}'$ is still at least 0.7.}

To proceed to a deeper analysis, we begin by examining node~2.  To obtain the surrogate inequality, we first take a linear combination of perturbed constraints 6, 7, and 9 with multipliers $(u_6,u_7,u_9)=(1,1,1)$.  Constraint~9 is perturbed by replacing the bound $0.3$ with $0.3+\Delta p$ to allow us to account for the effect of perturbations in this bound.  Constraint~6 is perturbed by replacing $t'$ with $t'+\Delta t'$ to permit an investigation of which lower bounds for $t'$ other than $0.7$ can be proved after the perturbation.  Since case (b) applies to node~2, we get the surrogate inequality $\delta' > -\Delta p - \Delta t'$.
This implies the single-literal clause $\delta'$ when $\Delta p=\Delta t'=0$, which means that $\delta'$ is the surrogate clause.  The surrogate inequality continues to imply clause $\delta'$ as long as 
\begin{equation}
\Delta p + \Delta t' \leq 0
\label{eq:post36}
\end{equation}
Node~3, which corresponds to case (c), similarly yields the surrogate inequality $-\delta'>-0.7 - \Delta t'$.  This implies the clause $\neg \delta'$ when $\Delta t'=0$, which is therefore the surrogate clause.  The surrogate inequality continues to imply $\neg \delta'$ as long as 
\begin{equation}
\Delta t' \leq 0.3
\label{eq:post37}
\end{equation}
The lower bound $0.7$ remains proven for $t'$ as long as the perturbation $\Delta p$ satisfies \eqref{eq:post36} and \eqref{eq:post37} with $\Delta t'=0$; that is, as long as $\Delta p \leq 0$, which is to say that $p$ (currently 0.3) can be reduced but not increased.  

If we wish to know under what conditions a lower bound of 0.9 (rather than 0.7) can be proved for $t'$, we can let $\Delta t'=0.2$ in \eqref{eq:post36} and \eqref{eq:post37}.  Since $\Delta t'=0.2$ satisfies \eqref{eq:post37}, condition \eqref{eq:post36} tells us that $\Delta p\leq -0.2$ suffices to prove $t'\geq 0.7+0.2$.  Thus if the upper bound on $v(P)$ is reduced from 0.3 to to 0.1, a lower bound of 0.9 can be proved for the truth value of $(P\supset Q)\supset\neg P$, regardless of the bound (if any) given for $v(Q)$.  As it happens, this is exactly the optimal value of $t'$ after the perturbation.  

The upshot of this analysis is that $(P\supset Q)\supset \neg P$ has a truth value of at least 0.7, given that $P$'s truth value is at most 0.3, regardless of the upper bound we place on $Q$'s truth value.  The guaranteed truth value 0.7 remains valid so long as the upper bound 0.3 on $P$'s truth value is not relaxed.  In fact, if it is tightened (reduced), the guaranteed truth value of $(P\supset Q)\supset \neg P$ increases by an equal amount, but if it is relaxed, the guaranteed truth value is reduced by an equal amount.  

The practicality of this type of comprehensive analysis depends on the number of leaf nodes that generate relevant surrogate inequalities.  The total of number of leaf nodes can be very large, but it is likely that only a fraction of them generate surrogate inequalities that contain perturbation terms for the constraints of interest.  The remainder of the leaf nodes can be ignored.  If we are interested only in learning which constraints play a role in deducing the optimal solution, we need only observe which constraints have a positive dual multiplier in at least one leaf node.  There is no need to generate and store surrogate inequalities.  This limited form of analysis can be performed on any problem instance that can be solved by a branch-and-bound method.

\subsection{MILP Postoptimality Analysis with Decision Diagrams}

A second method of postoptimality analysis for MILP, described in \cite{SerHoo20}, generates a BDD that represents near-optimal solutions of the problem. Such a BDD is convenient for conducting several types of postoptimality analysis.
One can efficiently build a BDD that compactly stores all such solutions during the normal branch-and-bound search for an optimal solution.  A simple procedure (``sound reduction'') allows it to be further compressed, often by several orders of magnitude, by introducing certain infeasible solutions that do not affect postoptimality analysis.  As a result, very large optimization problems can be analyzed in this fashion.

We illustrate the idea by applying it to the problem of inference in propositional logic.  Consider again the set $F=\{f_1,f_2,f_3\}$ of formulas listed in \eqref{eq:proj10}.  Suppose we wish to determine the minimum number of atomic propositions $x_i$ that can be false if $F$ is to be satisfied.  This can be ascertained by minimizing  $\sum_i (1-x_i)$ subject to $F$, a problem that can be solved by integer programming.  As it happens, at least two variables must be false.  We would like to know which of the formulas in $F$ play a role in this deduction.

\begin{figure}[!h]
	\centering
	\includegraphics[scale=0.95,clip=true,trim=10 10 10 10]{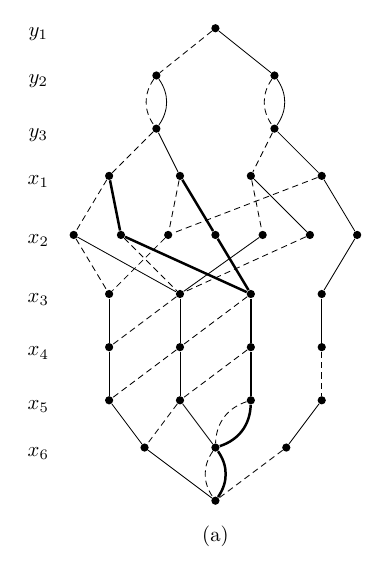} \hspace{4ex}
	\includegraphics[scale=0.95,clip=true,trim=10 10 10 10]{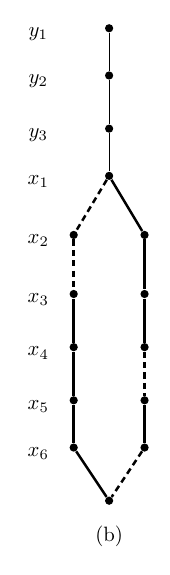} \hspace{4ex}
	\includegraphics[scale=0.95,clip=true,trim=10 10 10 10]{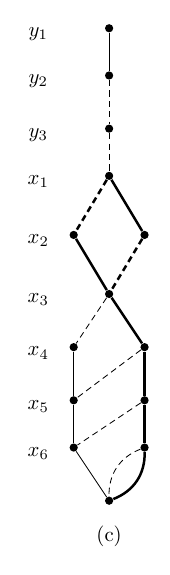} 
	\caption{(a) BDD for postoptimality analysis of inference from formulas \eqref{eq:proj10}.  (b) BDD after fixing $(y_1,y_2,y_3)=(1,1,1)$.  (c) BDD after fixing $(y_1,y_2,y_3)=(1,0)$.  Thick arcs indicate optimal values of $\bm{x}$.}
	\label{fig:BDDpostOptimality}
\end{figure}

This question can be addressed by introducing 0--1 variables $y_1,y_2,y_3$ that respectively indicate whether $f_1,f_2,f_3$ are enforced.  We augment $F$ by adding formulas $y_i\supset f_i$ for $i=1,2,3$ to obtain $F'$.  The minimum of $\sum_i(1-x_i)$ subject to $F'$ is of course zero, since the $y_i$s can all be 0, in which case no $f_i$ is enforced.  However, we can investigate the role of the $f_i$s by building a BDD that represents certain {\em near-optimal solutions} of this problem; namely, solutions with value 0, 1, or 2. Such a BDD appears in Fig.~\ref{fig:BDDpostOptimality}(a).

An optimal solution subject to $F'$ can be found by solving an easy shortest path problem in the BDD of Fig.~\ref{fig:BDDpostOptimality}(a).  Since we are minimizing the number of false $x_i$s, we place a cost (length) of 1 on dashed arcs in Fig,~\ref{fig:BDDpostOptimality} that correspond to values of the $x_i$s, and view all other arcs as having cost 0.  The shortest top-to-bottom path has zero cost, indicating that an optimal solution has value 0, as expected.  Now we can determine the result of enforcing any subset of $f_i$s by fixing the corresponding $y_i$s to 1 and computing shortest paths in the simplified BDD that results. 

We can see right away that $f_2$ has no bearing on the problem, since the value of $y_2$ has no effect on any path length.  If we enforce all the $f_i$s by fixing all $y_i$s to 1, we get the BDD in Fig.~\ref{fig:BDDpostOptimality}(b), in which the shortest path length is 2, as expected.  If we enforce only $f_1$, we obtain the BDD of Fig.~\ref{fig:BDDpostOptimality}(c), in which the shortest path has length 1.  Thus without premise $f_3$, the minimum number of false atoms drops to 1.  It is similarly shown
 that without $f_1$, the number drops to 0.  
 
The BDDs also allow inferences regarding possible values of the variables $x_i$.  If all three formulas are enforced, as in Fig.~\ref{fig:BDDpostOptimality}(b), we see immediately that the number of false atoms can be reduced to 2 only by setting $x_3$ and $x_5$ to true.  
If we enforce only $f_1$, as in  Fig.~\ref{fig:BDDpostOptimality}(c), we see that the number of false atoms can be reduced to 1 only by setting $x_1$ or $x_2$ to false.

\section{Answer Set Programming Modulo Theories} \label{sec:answerSet}

Optimization can play an important role when logical and arithmetical constraints appear in the same problem.  This occurs, for example, when a propositional satisfiability (SAT) problem is enhanced with constraints involving integers or real numbers, resulting in a {\em SAT modulo theory} (SMT) \cite{BarTin18,TanWanHe24}.  The term refers to the fact that the logical propositions are evaluated in the context of (``modulo'') a specific mathematical theory, such as linear equations or inequalities.  

SMTs are not primarily identified with AI, since they have many applications in other areas, such as hardware and software verification and general combinatorial problem solving.  However, a similar problem class based on {\em answer set programming} (ASP) is more closely related to AI, due to its use of default logic rather than a traditional SAT problem \cite{ErdGelLeo16,JanNie16,Lif99,Lif08,MarTru99,Nie99}. Like a SAT problem, a default logic problem can be combined with arithmetical or other types of constraints, resulting in an {\em ASP modulo theory} \cite{GebKamKauOstSchWan16,JanKamOstSchWanSch17,LeeMen13}.  Some applications of ASP modulo theories are described in \cite{BarLee14,LeeMen13,Lie23,VarMen24,WalSchBha17}.  We show how an optimization method such as linear or integer programming can play a key role when arithmetical constraints are involved.

ASP modulo theories (as well as SMTs) are closely related to logic-based Benders decomposition (LBBD), an optimization method we applied to projection problems in Section~\ref{sec:projection}.  The default logic problem can be regarded as a master problem with propositional variables $x_i$.  The subproblem can be a numerical constraint set whose feasibility can be checked with a linear or integer programming solver.  For example, the subproblem may consist of linear inequalities that are activated by a model $\{x_i\;|\;i\in I\}$ obtained when solving the master problem.  That is, the subproblem may consist of constraints of the form $x_i\rightarrow \bm{a}^i\bm{y}\leq b_i$, where $\bm{y}\geq\bm{0}$ is a vector of real-valued variables.  An infeasible subproblem generates a Benders cut in the form of a nogood constraint $\bigvee_{i\in I'} \neg x_i$, where $I'$ indexes a subset of the model that is responsible for the infeasibility.  A stable model can be found by repeatedly using LBBD to find minimal models of the combined logical and inequality constraints, until a model is found that is also minimal in the corresponding Gelfond-Lifschitz (G-L) transform of the logical constraints.  

As an example, consider the default logic problem \eqref{eq:mono10} with the following additional constraints
\begin{equation}
\begin{array}{l}
x_1 \rightarrow 3y_1 + 2y_2 \leq 6 \\
x_2 \rightarrow -3y_1+y_2 \leq -9 \\
x_3 \rightarrow -2y_1 -5y_2 \leq -10 \\
y_1, y_2 \geq 0  
\end{array} \label{eq:answer01}
\end{equation}
The goal is to identify a stable model, which can be found by enumerating minimal models of the constraints in \eqref{eq:mono10} and \eqref{eq:answer01}, until a model is found that is also minimal in the G-L transform of \eqref{eq:mono10}.  If no such model is found, the problem has no stable model.

To find a minimal model of the constraints in \eqref{eq:mono10} and \eqref{eq:answer01}, we first use integer programming (as in Section~\ref{sec:default}) to find a minimal model of the master problem \eqref{eq:mono10}.  We then use linear programming (LP) to check whether the inequalities in the subproblem  \eqref{eq:answer01} that are enforced by this model can be satisfied.  If not, we consult the dual solution of the LP to generate a Benders cut in the form of a nogood constraint that excludes the infeasible model.  We then generate another minimal model by solving the master problem subject to the nogood constraint.  We continue in this fashion, generating nogood constraints as needed, until finding a minimal model that satisfies \eqref{eq:answer01}, or until the master problem is infeasible. 

The entire process in detail goes as follows.  We begin as in Section~\ref{sec:default} by solving the integer programming problem \eqref{eq:mono12} to find a minimal model $\{x_1,x_2\}$ of \eqref{eq:mono10}.  This model violates \eqref{eq:answer01} because it enforces the first two inequalites in \eqref{eq:answer01}, and an LP solver determines that they cannot be satisfied when $y_1,y_2\geq 0$.  If we associate dual variables $u_1,u_2$ with these inequalities, the LP solver yields a dual solution $(u_1,u_2)=(1,1)$ that proves infeasibility.  Since both dual variables are positive, we know that $x_1$ and $x_2$ are jointly responsible for the infeasibility.\footnote{In general, the subset $S$ of constraints corresponding to positive dual variables is infeasible.  However, proper subsets of $S$ may also be infeasible, and they would give rise to stronger nogood constraints.  The problem of finding an irreducible infeasible subset is discussed in \cite{Chi96,Chi97,Chi01,ChiDra91}.}  We therefore generate a nogood clause $\neg x_1 \vee \neg x_2$ to ensure that they do not both occur in a model.  
(If only $u_1$ were positive, we would generate a nogood clause consisting of the single literal $\neg x_1$.)    
We add the nogood clause (in the form $x_1+x_2\leq 1$) to the integer programming problem \eqref{eq:mono12}, which yields an optimal solution $(x_1,\dots,x_5)=(0,0,0,1,1)$.   This solution vacuously satisfies \eqref{eq:answer01} because none of the inequalities in \eqref{eq:answer01} are enforced.  However, the corresponding model $\{x_4,x_5\}$ is not minimal for the G-L transform, which consists of the single proposition $(x_1\wedge x_4)\rightarrow x_5$.  

At this point we attempt to find another minimal model for the constraints in \eqref{eq:mono10} and \eqref{eq:answer01} by solving the integer programming problem \eqref{eq:mono12} with an additional constraint $x_4+x_5\leq 1$ to exclude the model found previously.  We also retain the nogood constraint $x_1+x_2\leq 1$ in \eqref{eq:mono12} because it remains valid.  The resulting solution $(x_1,\ldots,x_5)=(1,0,1,0,1)$ satisfies \eqref{eq:answer01}, but the  corresponding model $\{x_1,x_3,x_5\}$ is not minimal in the G-L transform, which consists of $x_1\rightarrow x_3$, $(x_1\wedge x_4)\rightarrow x_5$, and $\rightarrow x_1$.  The integer programming problem \eqref{eq:mono12} becomes infeasible after adding a constraint $x_1+x_3+x_5\leq 2$ to exclude this last solution.  We conclude that the ASP modulo theory problem consisting of \eqref{eq:mono10} and \eqref{eq:answer01} has no stable model.

\section{Possible Reserch Directions}

Many opportunities remain for application of optimization methods to logical inference problems related to AI.  One avenue is to exploit the mathematical properties of semantics for various logics.  Boolean, probabilistic, belief, and nonmonotonic logics have already been discussed in the preceding pages.  Other possibilities include Kripke semantics for modal logics involving necessity and possibility (S4 and S5), as well as various semantics for epistemic, doxastic and deontic logics \cite{BladeRijVen02,BlavanBen06,GioSai26,Gol03,GorOtt06,vanDitHalHoeKoo15}.

There are, in fact, deep connections between logical semantics and optimization theory that can serve as the basis for applying one to the other \cite{BorChaMit02,ChaHoo91,ChaHoo99,ConCor94,Jer89,Wil09}.  Even traditional Boolean logic has a rich polyhedral structure when truth assignments are interpreted as vertices of a hypercube, which opens the door to specialized integer programming methods \cite{ChaHoo99}.  The resolution method of propositional logic underlies the fundamental theorem of cutting plane theory in integer programming \cite{Chv73}, and resolvents are special cases of the Chv\'{a}tal-Gomory cuts widely used in integer programming solvers \cite{ConCorZam14,Hoo89,NemWol99}.  
An optimization-oriented analysis of semantics that have been developed for various modal logics could lead to new and efficient inference methods.

Still other possible research directions arise in the area of Bayesian and causal networks.  Optimization methods have already been applied to learning the structure of traditional Bayesian networks \cite{BarCus17,KitConGuoLiuCho23} as well as to maximum a posteriori (MAP) inference \cite{DubLeeFla25,SonMelGloJaaWei08}.  One could investigate extending these methods to networks that allow Boolean formulas in the manner of Bayesian logic (Section~\ref{sec:BayesianLogic}).  In addition, the large literature on causal networks \cite{Pea95,Pea00,WeiPreFar25,ZanOzkSte22} may offer further opportunities.  It may be possible, for example, to specialize inference methods for Bayesian logic to networks with causal structure, as well as to develop an analog of do-calculus for Bayesian logic.  One might also adapt methods for optimizing causal interventions \cite{VonVerdeNobelBraMalBacLaaKon24} to Bayesian logic.  Another possibility is to exploit the structure of small-treewidth Bayesian logic problems.  This has long been done for traditional Bayesian networks \cite{KorPar13,LauSpi88,PeyCrodeGivFraRobSabSchVig18}, and as remarked in Section~\ref{sec:projectionBDD}, small treewidth can also be exploited in logical inference.   The two approaches might be combined in the context of Bayesian logic, perhaps through nonserial dynamic programming, a concept known in the optimization community since the 1970s \cite{BerBri72}.    

An intriguing possibility is to harness the full potential of projection, which we have seen is a generalization of both optimization and logical inference.  By projecting a knowledge base onto a subset of Boolean variables, one could not only answer queries as described in Section~\ref{sec:projection}, but create smaller, specialized rule bases for efficient application in particular domains.  It was noted in Section~\ref{sec:projection.Benders} that a projection onto discrete variables can be computed by running Benders decomposition to completion.  The linear programming models of belief logics require polyhedral projection, but it is a well-studied (if not particularly well-solved) problem in optimization theory  \cite{Bal98,HowKin12,HuyLasLas92,JinMorTal20,JonKerMac08}.  An interesting question is how to combine polyhedral projection with column generation for application to Boole's probabilistic logic.  Finally, there is the possibility of representing a domain-specific projected knowledge base in the form of a decision diagram, in order to obtain enhanced transparency and enable rapid responses to complicated queries.

%


\end{document}

%% file: extra_header_changes_Marty.tex
\theoremstyle{definition}

\def\emptyset{\varnothing}